\documentclass{article}
\usepackage{iclr2027_conference,times}
\usepackage[hyphens]{url}
\usepackage{graphicx}
\usepackage{booktabs}
\usepackage{float}
\usepackage{multirow}
\usepackage{amsmath}
\usepackage{array}
\usepackage{tabularx}
\usepackage{tcolorbox}
\tcbuselibrary{breakable,skins}
\usepackage[hidelinks]{hyperref}

\newcolumntype{Y}{>{\centering\arraybackslash}X}
\newtcolorbox{promptbox}[1][]{
  breakable,
  colback=blue!2,
  colframe=black!65,
  colbacktitle=blue!8,
  coltitle=black,
  fonttitle=\bfseries,
  title=#1,
  boxrule=0.5pt,
  arc=1pt,
  left=5pt,
  right=5pt,
  top=5pt,
  bottom=5pt,
  before skip=6pt,
  after skip=6pt
}
\newcolumntype{Z}[1]{>{\centering\arraybackslash}m{#1}}

\title{Beyond Captions: Context-Grounded Reconstruction for Biomedical Multimodal Continued Pretraining}

\author{
  \textbf{Guanghao Zhu$^{1}$ \enspace
  Zeyu Liu$^{1}$ \enspace
  Zhitian Hou$^{1}$ \enspace
  Pengkai Wang$^{1}$ \enspace
  Zhijie Sang$^{2}$ \enspace
  Yang Yu$^{1}$} \\
  \textbf{Minheng Ni$^{1}$ \quad
  Wenjun Wang$^{1}$ \quad
  Yanggan Gu$^{1}$ \quad
  Shuo Cai$^{1}$ \quad
  Congkai Xie$^{2}$} \\
  \textbf{Jianmin Wu$^{1,3}$ \quad
  Hongxia Yang$^{1,2,3}$}\thanks{Corresponding author. Email: \texttt{hongxia.yang@polyu.edu.hk}.} \\
  \normalfont
  $^{1}$The Hong Kong Polytechnic University \\
  $^{2}$InfiX.ai \\
  $^{3}$Research Institute of Generative Artificial Intelligence
}

\iclrfinalcopy

\begin{document}
\maketitle
\lhead{Preprint}

\begin{abstract}
  Biomedical figures are explained not by captions alone but by body-text passages that discuss them. Yet current multimodal corpora typically reduce figures to isolated image-caption pairs, discarding this crucial context. Existing pipelines either omit this context or append it without enforcing the figure references that support each attachment, which can create unsupported image-text attachments and incoherent discourse. We introduce context-grounded reconstruction, a source-grounded framework that converts PubMed Central Open Access (PMC-OA) records into referentially coherent interleaved sequences. It recovers captions and source text, attaches context only through article-native figure references, repairs non-contiguous context, and prunes unsupported images. Starting from these reconstructed sequences, PMC-InterCPT first filters records for text quality and medical relevance, then applies evidence-aware allocation to form a 9.63B-token corpus for continued pretraining (CPT) of generative medical MLLMs. With fixed supervised fine-tuning (SFT), PMC-InterCPT improves Qwen3.5-4B-Base by 1.46 medical-average points and 3.11 general/scientific-average points over a token-matched raw source control, and surpasses a 42\% larger raw-data run. Gains transfer to Qwen3.5-2B-Base and LLaVA-OneVision-1.5-4B-Base. Controlled ablations show that context-grounded reconstruction, rather than simply appending article context, is central to useful biomedical multimodal CPT.
\end{abstract}

\section{Introduction}
Multimodal large language models (MLLMs) benefit from interleaved image-text pretraining that preserves rich visual and textual evidence in a common sequence~\citep{lin2024vila}. In biomedical literature, that evidence is rarely contained in a caption alone. The surrounding body text can identify cohorts, experimental settings, comparisons, and claims that make a figure interpretable. It also frequently refers to several figures jointly. A biomedical CPT corpus that retains this structure could provide substantially richer training signal than an image-caption archive.

Existing resources extracted from PubMed Central Open Access (PMC-OA), including BIOMEDICA~\citep{lozano2025biomedica}, expose biomedical figures at unprecedented scale. BIOMEDICA contains more than 24 million image-caption pairs and extensive metadata, but its primary training representation is a figure paired with a caption. This organization leaves figure-referencing article text unused. More importantly, simply concatenating nearby text is unsafe: a paragraph can mention multiple figures, occur far from other retained contexts, or remain after the figure it supports has been removed. We therefore formulate the task as \emph{grounded reconstruction}: rebuilding article-native records into referentially coherent interleaved sequences while preserving the figure references that justify each image-context association.

We propose context-grounded reconstruction, a reference-constrained framework for constructing biomedical CPT corpora. Rather than treating article text as generic auxiliary text, it uses article-native figure references as structural evidence for context attachment. The framework begins with source-grounded recovery, which restores provenance-linked captions and normalizes source text while preserving reference anchors. Reference-constrained interleaving then forms multi-figure sequences without duplicating shared context. Coherence-constrained curation detects discontinuities among retained source passages, and preserves locally supported context. These operations yield a candidate pool in which each retained image--caption--context relation remains traceable to the source article. Semantic curation filters this pool for textual usability and medical relevance at full-corpus scale. Finally, evidence-aware allocation shapes the CPT learning signal by regulating the model's exposure to complementary forms of scientific evidence: biomedical visual observations, quantitative results, mechanistic structures, and auxiliary context.

We instantiate this process as PMC-InterCPT and evaluate it as CPT data for generative medical MLLMs under a fixed SFT stage. Rather than validating the corpus only through CLIP-style contrastive representation scores, we assess its effect on generative medical VQA, medical reasoning, and scientific figure understanding, using token-matched raw controls and controlled ablations.

Our main contributions are:
\begin{itemize}
  \item \textbf{Reference-constrained interleaving}, a reconstruction formulation that converts PMC figure records and article text into referentially coherent interleaved sequences, including multi-figure relations and context-supported image retention.
  \item \textbf{Coherence-constrained curation}, a procedure that turns reconstructed sequences into usable biomedical CPT data. It repairs context, prunes unsupported images, and filters for text quality and medical relevance.
  \item \textbf{Evidence-aware allocation}, a four-bucket taxonomy and controlled allocation scheme that regulates the model's exposure to clinical visual, quantitative, and mechanistic scientific evidence. PMC-InterCPT is the resulting 9.63B-token corpus of this allocation.
  \item \textbf{Generative MLLM gains.} With fixed SFT, PMC-InterCPT outperforms the token-matched raw baseline by 1.46 points on medical tasks and 3.11 points on general/scientific tasks, with gains transferring across backbones.
\end{itemize}

\section{Related Work}
\subsection{Medical Multimodal Large Language Models}
Recent years have witnessed rapid development of medical multimodal large language models (MLLMs), including LLaVA-Med~\citep{li2023llava}, Med-Flamingo~\citep{moor2023med}, HuatuoGPT-Vision~\citep{chen2024huatuogptvision}, MAIRA-2~\citep{bannur2024maira}, and LingShu~\citep{xu2025lingshu}, which adapt general-purpose VLMs to clinical and biomedical domains through instruction tuning or domain-adaptive pretraining.
Despite strong performance on medical VQA tasks, these models typically rely on small curated instruction datasets or general-domain pretrained weights, without access to large-scale high-quality medical interleaved CPT data.
Our work addresses this gap by constructing and validating a CPT dataset specifically designed for this purpose.

\subsection{Biomedical Multimodal Datasets from Literature}
PMC-OA~\citep{lin2023pmcclip}, PMC-15M~\citep{zhang2023biomedclip}, Open-PMC-18M~\citep{baghbanzadeh2025open}, and BIOMEDICA~\citep{lozano2025biomedica} demonstrate the value of PMC as a scalable source of biomedical image-caption data. These resources differ in scale, subfigure handling, and metadata, but their predominant training unit is an image-caption pair. Their reported empirical validation focuses on CLIP-style contrastive medical vision-language representation learning, rather than evaluating whether the resulting data improves generative medical MLLMs. PubMedVision reformats PubMed image-text pairs into medical VQA data using an MLLM~\citep{chen2024huatuogptvision}, which is tailored to instruction-style supervision rather than CPT. Open-PMC-18M additionally shows that curation and alignment can matter more than raw scale~\citep{baghbanzadeh2025open}.

PMC-InterCPT is complementary to these efforts. Rather than treating adjacent article text as generic enrichment or converting records into question-answer pairs, we reconstruct figure-reference-supported interleaved sequences and explicitly repair incoherent context before CPT.

\subsection{Interleaved Image-Text Pretraining}
Flamingo~\citep{alayrac2022flamingo} demonstrated that interleaved image-text data enables strong few-shot multimodal generalization.
Subsequent work including MMC4~\citep{zhu2023multimodal} and OBELICS~\citep{laurenccon2023obelics} scaled interleaved web data to billions of documents, showing consistent benefits over image-caption pair training.
These results support interleaved biomedical representations but leave open how article context should be grounded to figures. We address this through reference- and discourse-constrained reconstruction from PMC.

\subsection{Data Quality and Mixture for Pretraining}
FineWeb~\citep{penedo2024fineweb}, DCLM~\citep{li2024datacomp}, and RefinedWeb~\citep{penedo2023refinedweb} have established that careful quality filtering of web text yields substantial gains for general LLM pretraining.
DoReMi~\citep{xie2023doremi} showed that domain reweighting can significantly accelerate pretraining convergence.
We extend these principles to medical multimodal CPT through a corpus construction process that begins with source-grounded reconstruction, filters records with LLM-supervised quality models, and then allocates the retained data with a modality-aware resampling strategy. We provide a systematic empirical study of these stages.

\section{Method}
\begin{figure}[t]
  \centering
  \includegraphics[width=\textwidth]{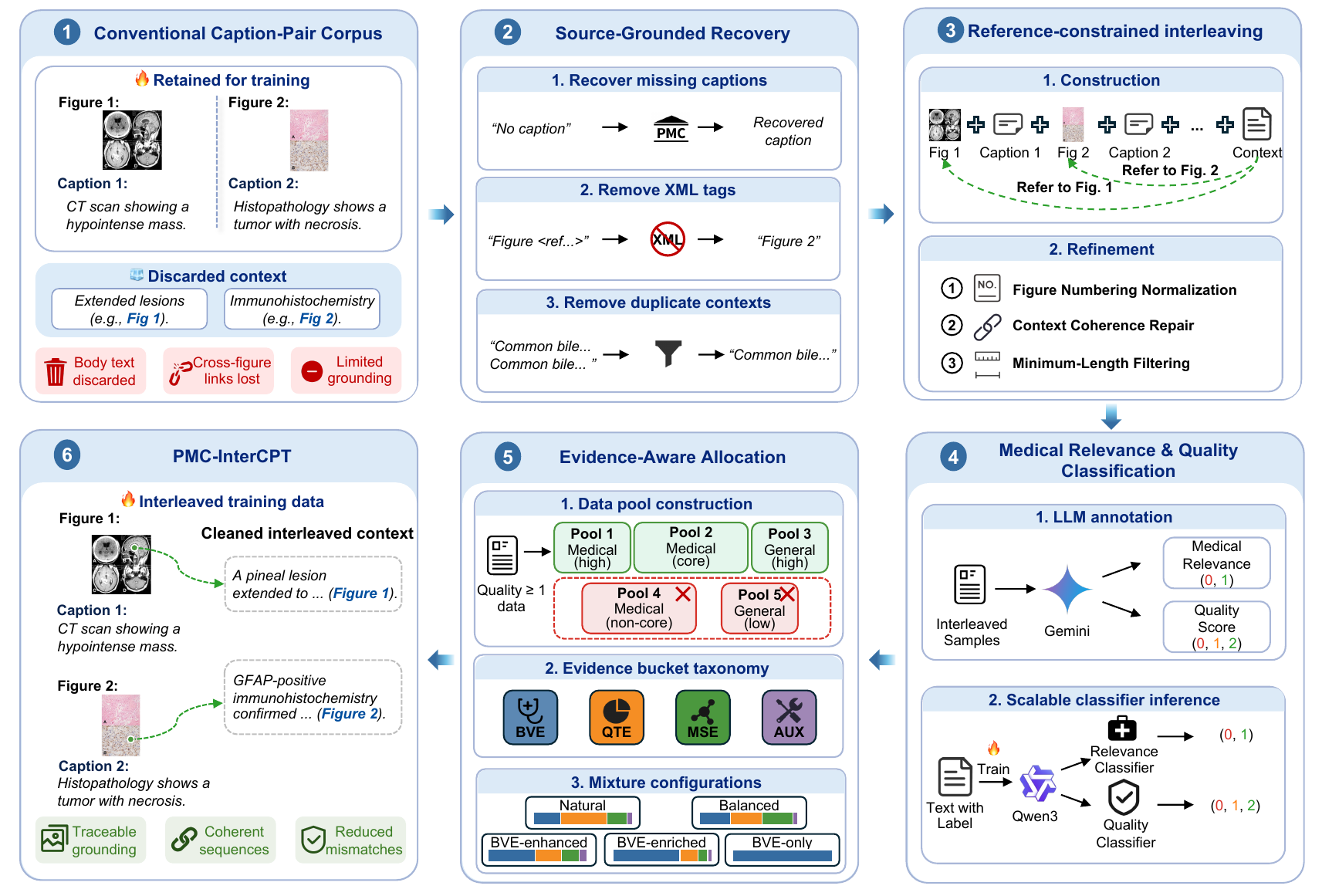}
  \caption{From conventional caption pairs to PMC-InterCPT. Source-grounded recovery restores missing captions and cleans article text. Reference-constrained interleaving attaches and refines passages linked by article-native figure references. Medical relevance and text quality guide source-pool selection, and evidence-aware allocation forms the final interleaved corpus.}
  \label{fig:pipeline}
\end{figure}

BIOMEDICA is a large-scale archive of biomedical image-caption pairs extracted from the PMC Open Access subset. Its article-level provenance makes it possible to reconstruct a stronger training unit: a sequence in which each retained image and context segment is justified by an article-native figure reference. As Figure~\ref{fig:pipeline} illustrates, conventional caption-pair training retains figures and captions but discards the source article context that explicitly links them. We call the process that recovers this structure \emph{context-grounded reconstruction}. It addresses missing source text, flattened markup and repetition, many-to-many figure-context references, and incoherent concatenation of distant paragraphs. The reconstructed sequences are subsequently curated for quality and medical relevance, then allocated across complementary evidence types to form the final CPT corpus.

\subsection{Source-Grounded Recovery}

\paragraph{Caption Recovery.} BIOMEDICA can contain records without a usable caption when the original PMC-OA extraction does not associate an image or table-derived media object with a directly recoverable caption field. For such records, we recover the caption directly from the corresponding PMC XML by mapping the image to its enclosing \texttt{<fig>} or \texttt{<table-wrap>} element. We retain the source figure identifier and use identifier variants as a fallback. This recovery is preferable to generating a caption because it preserves the publication's original figure-text association.

\paragraph{Structure-preserving normalization.} We remove residual XML tags from both captions and context fields while retaining their inner text. We then normalize whitespace, remove local near-duplicate fragments, and deduplicate repeated contexts within an article. These operations correct artifacts introduced when XML paragraphs with inline elements are flattened, while retaining the references needed for subsequent reconstruction.

\subsection{Reference-Constrained Interleaving}

Each reconstructed record alternates image slots with caption and context slots. A context paragraph is attached only to figures that it explicitly references. When a paragraph refers to multiple figures, the corresponding figures are co-included rather than duplicating the paragraph across independent examples. We canonicalize caption prefixes with source figure numbers, use each context only once within an article, and propagate image modality metadata for mixture allocation.

Context segments for a figure can originate from distant article locations. Concatenating them can place unrelated statements next to one another, producing a synthetic context that may not faithfully represent the evidence associated with the figure. We therefore retain adjacent sequences and resolve a discontinuity by favoring a segment exclusively referring to the primary figure; otherwise, we retain the earlier segment. Any non-primary image no longer supported by the retained context is pruned, while the primary image is always preserved. We finally remove examples that do not contain sufficient caption and context for meaningful CPT signal using separate minimum-length criteria for captions and contexts. Exact thresholds are reported in the appendix.

\subsection{Medical Relevance and Quality Classification}

We filter reconstructed samples by medical relevance and text quality using a two-stage approach.

\paragraph{Stage 1: Annotation.}
We use Gemini-3.1-pro-preview to assign medical relevance $\in \{0,1\}$ and text quality $\in \{0,1,2\}$. The annotator receives captions and figure-linked context but not images. Original PMC-OA figure-caption pairings and explicit figure references provide structural evidence for these associations.

We do not use VLM-based image-text relevance as a hard filter. In a paired audit, image-conditioned scoring gave unsupported reasons for penalties or missed explicit mismatches (Appendix~\ref{app:teacher-quality-audit}). The quality label instead targets malformed table or \LaTeX{} dumps, repetition, and incoherent text. Text-only inference also avoids processing image tokens at corpus scale. Sampled labels are reviewed before classifier training.

\paragraph{Stage 2: Classifier Inference.}
We use the reviewed annotations to train separate Qwen3-1.7B classifiers for medical relevance and text quality.
The medical classifier predicts $m \in \{0,1\}$, where $m=1$ denotes medical relevance, while the quality classifier predicts $q \in \{0,1,2\}$.
We apply these predicted labels to every reconstructed record and retain $q\geq1$ records for source-pool construction. A quality score of $q=0$ denotes severely corrupted text, such as heavy repetition, malformed table or \LaTeX{} dumps, or incoherent passages, and is removed. Scores $q=1$ and $q=2$ denote usable-but-imperfect and clean, coherent text, respectively. Annotation prompts, classifier training details, and per-class results are provided in the appendix.

\subsection{Evidence-Aware Allocation}
\label{sec:resampling}

\paragraph{Source Pool.}
We exclude all records from PMC articles represented in the PMC-VQA-clean evaluation split. We then partition the remaining quality$\geq$1 data into five pools based on quality score, medical relevance, and image modality:
(1)~\emph{high-quality medical} (quality=2, medical=1);
(2)~\emph{medical strong-related supplement} (quality=1, medical=1, label $\in$ \{Clinical Imaging, Immuno Assays, Laboratory Specimens and Cultures, Microscopy, PCR\});
(3)~\emph{high-quality general} (quality=2, medical=0);
(4)~\emph{medical weak-related supplement} (quality=1, medical=1, non-core labels); and
(5)~\emph{low-quality general} (quality=1, medical=0).
Pools 4 and 5 are excluded from pretraining: pool 4 contains lower-quality medical samples from weakly related visual categories, while pool 5 contributes off-domain content.
The source pool for all subsequent experiments therefore consists of pools 1--3: high-quality medical, medical strong-related supplement, and high-quality general.

\paragraph{Evidence Buckets.}
Analysis of the natural token distribution within the source pool reveals severe imbalance: Plots and Charts account for 51.2\% of tokens, while clinically relevant Clinical Imaging represents only 7.5\%.
To enable principled resampling, we define four \emph{evidence buckets} that group image modalities by their functional role in biomedical communication:

\begin{itemize}
  \item \textbf{Biomedical Visual Evidence (BVE):} Clinical Imaging, Microscopy, Immuno Assays, PCR, Laboratory Specimens and Cultures
  \item \textbf{Quantitative/Table Evidence (QTE):} Plots and Charts, Tables
  \item \textbf{Mechanism/Structure Evidence (MSE):} Illustrative Diagrams, Chemical Structures, Graphs and Networks, Scientific Formulae and Equations
  \item \textbf{Auxiliary (AUX):} Natural Images, Tools and Materials, Screen Based Visuals, Maps, Ambiguous
\end{itemize}

\paragraph{Allocation Strategy.}
Evidence-aware allocation controls the relative exposure to complementary forms of scientific evidence in the CPT corpus. All allocations draw from the same filtered source pool. The allocation sets the token weights across buckets, and within each bucket, samples are drawn in proportion to their natural token count. We compare five allocations at a 1B-token CPT budget (Section~\ref{sec:mixture}). Natural preserves the source-pool distribution, Balanced assigns more even weights to the evidence buckets, BVE-enhanced targets 45\% BVE, BVE-enriched targets 70\% BVE while preserving the relative weights of QTE, MSE, and AUX from BVE-enhanced, and BVE-only assigns the full budget to BVE. The final 9.63B-token PMC-InterCPT corpus uses BVE-enhanced (45\% BVE, 30\% QTE, 20\% MSE, and 5\% AUX), which achieved the highest medical average among the five tested allocations and retained enough non-BVE evidence to construct the full corpus from the source pool.

\subsection{Dataset Summary}
\label{sec:dataset-statistics}

\begin{figure}[t]
  \centering
  \includegraphics[width=\textwidth]{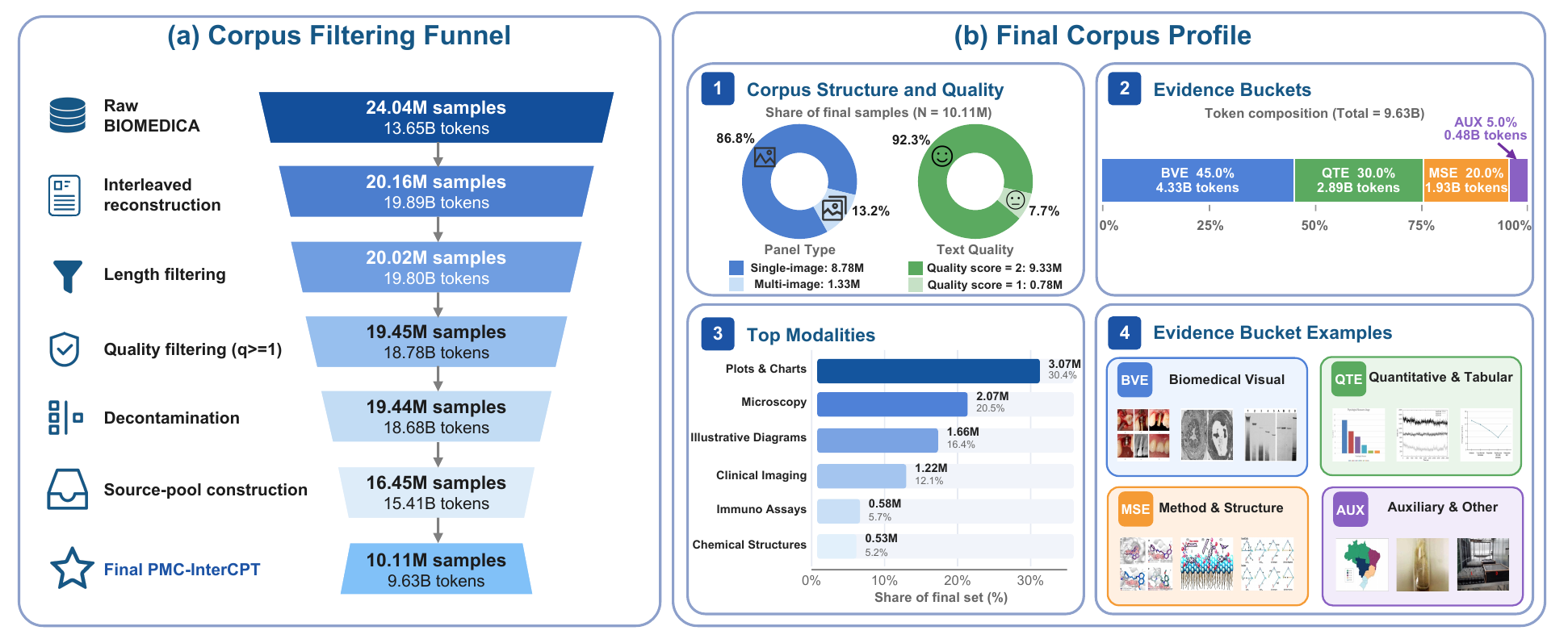}
  %BEGIN DIFNOMARKUP
  \par\nointerlineskip
  %END DIFNOMARKUP
  \caption{Corpus filtering funnel and final profile of PMC-InterCPT. (a) The funnel tracks sample and token counts from raw BIOMEDICA through interleaved reconstruction, length filtering, quality filtering, decontamination, source-pool construction, and final allocation. (b) The final profile reports panel structure and text quality over 10.11M samples, top image modalities by sample share, and the BVE-enhanced token allocation. Representative images illustrate each evidence bucket.}
  \label{fig:statistics}
\end{figure}

Figure~\ref{fig:statistics} summarizes both the scale changes and the final composition of PMC-InterCPT. Adding figure-linked context expands the raw 13.65B-token caption archive into a 19.89B-token interleaved pool. Length filtering, quality curation, decontamination, source-pool construction, and evidence-aware allocation then yield the final corpus of 10.11M samples and 9.63B tokens. This contraction reflects selective construction of context-grounded training sequences rather than raw-data scaling.

The final corpus remains structurally diverse: 13.2\% of samples contain multiple images, preserving cross-figure evidence unavailable to caption-only training. Plots and charts are the largest individual modality (30.4\% of samples), followed by microscopy (20.5\%) and illustrative diagrams (16.4\%). At the allocation, BVE receives the largest share of tokens, while QTE and MSE together account for half of the corpus. The representative image panels in Figure~\ref{fig:statistics}(b) illustrate this coverage across the four evidence buckets.

\section{Experiments}
\subsection{Experimental Setup}

\paragraph{Model.}
Our primary controlled data study uses Qwen3.5-4B-Base~\citep{qwen2026qwen35} before a fixed SFT stage, so differences primarily reflect the CPT data source and mixture. We additionally report Qwen3.5-2B-Base results at the same full scale and a separate architecture-transfer ablation on LLaVA-OneVision-1.5-4B-Base.

\paragraph{Evaluation.}
We evaluate on five medical benchmarks: MMMU-Med-Test~\citep{yue2024mmmu}, MMMU-Pro-Med-10~\citep{yue2025mmmu}, PMC-VQA-clean~\citep{zhang2023pmcvqa}, OmniMedVQA~\citep{hu2024omnimedvqa}, and PretexEval~\citep{zhou2024pretexeval}. We additionally evaluate on four general or scientific multimodal benchmarks: MMMU-All-Val~\citep{yue2024mmmu}, ChartQA~\citep{masry2022chartqa}, CharXiv-Val~\citep{wang2024charxiv}, and SciVQA~\citep{borisova2025scivqa}. The suite is intended to evaluate generative MLLM utility beyond contrastive representation learning, spanning medical visual question answering, medical factual knowledge, and scientific figure understanding. ChartQA, CharXiv-Val, and SciVQA are scored using Qwen2.5-72B as an LLM judge, and other benchmarks are scored by rule-based extraction. We found no exact image matches between the final PMC-InterCPT WebDataset and any visual benchmark in this main evaluation suite. We also compared the arXiv source papers and figures for CharXiv-Val and SciVQA-Test with the final CPT corpus and found no overlapping papers or matching figures.

\paragraph{Training protocol.}
All experiments use a two-stage CPT+SFT protocol.
During CPT, we use a maximum sequence length of 8192 with sequence packing, freeze the ViT encoder, and train for one epoch with a warmup ratio of 0.05 to a maximum learning rate of $5.0\times10^{-6}$, followed by cosine decay to $1.0\times10^{-6}$, micro-batch size 2, and global batch size 256.
CPT is run on 64 NVIDIA H800 GPUs.
After CPT, all models are supervised fine-tuned on the same 50K examples randomly sampled from LLaVA-OneVision~\citep{li2024llavaonevision}.
During SFT, we train all model parameters for three epochs on 8 NVIDIA H800 GPUs with cosine learning-rate decay from $2.5\times10^{-6}$ to 0, a warmup ratio of 0.1, micro-batch size 2, and global batch size 64.

\subsection{Main Results}
\label{sec:main-results}

Table~\ref{tab:main-results} compares PMC-InterCPT with both full raw BIOMEDICA and a 9.63B-token random BIOMEDICA control under the same fixed SFT stage, across Qwen3.5-4B-Base and Qwen3.5-2B-Base. The token-matched control isolates the effect of reconstruction from CPT token budget.

\begin{table}[t]
  \caption{Full-scale CPT+SFT results across Qwen3.5 Base scales. Superscripts give CPT token budgets. Every row uses the same 50K-example LLaVA-OneVision SFT stage. Bold and underlining mark the best and second-best values.}
  \label{tab:main-results}
  \vspace{3pt}
  \centering
  {\fontsize{6}{7}\selectfont
    \setlength{\tabcolsep}{1.0pt}
    \begin{tabular*}{\textwidth}{@{\extracolsep{\fill}}cc*{11}{c}}
      \toprule
      \textbf{Backbone} & \textbf{CPT data} & \multicolumn{6}{c}{\textbf{Medical benchmarks}} & \multicolumn{5}{c}{\textbf{General/scientific benchmarks}} \\
      \cmidrule(lr){3-8}\cmidrule(lr){9-13}
      & & \textbf{\shortstack{MMMU\\Med}} & \textbf{\shortstack{MMMU-Pro\\Med}} & \textbf{\shortstack{PMC\\VQA}} & \textbf{\shortstack{Omni\\Med}} & \textbf{\shortstack{Pretex\\Eval}} & \textbf{\shortstack{Medical\\avg.}} & \textbf{\shortstack{MMMU\\All}} & \textbf{\shortstack{Chart\\QA}} & \textbf{\shortstack{Char\\Xiv}} & \textbf{\shortstack{Sci\\VQA}} & \textbf{\shortstack{General\\avg.}} \\
      \midrule
      \multirow{4}{*}{Qwen3.5-4B} & None (SFT only) & \textbf{58.96} & \textbf{41.61} & 61.40 & 85.37 & 64.52 & 62.37 & \textbf{55.11} & \underline{75.68} & \underline{53.50} & \textbf{68.71} & \underline{63.25} \\
      & BIOMEDICA$^{13.65\mathrm{B}}$ & 58.62 & 37.76 & \textbf{62.90} & 84.97 & 67.48 & 62.35 & 53.89 & 75.00 & 47.90 & 66.74 & 60.88 \\
      & BIOMEDICA$^{9.63\mathrm{B}}$ & 56.39 & 39.51 & 61.80 & \textbf{86.95} & \underline{67.68} & \underline{62.47} & 54.33 & 74.56 & 45.20 & 66.76 & 60.21 \\
      & PMC-InterCPT$^{9.63\mathrm{B}}$ & \underline{58.73} & \underline{40.56} & \underline{62.75} & \underline{86.51} & \textbf{71.10} & \textbf{63.93} & \underline{55.00} & \textbf{77.00} & \textbf{54.00} & \underline{67.29} & \textbf{63.32} \\
      \midrule
      \multirow{4}{*}{Qwen3.5-2B} & None (SFT only) & \underline{45.78} & \textbf{32.17} & 56.45 & \underline{80.35} & \underline{56.15} & \underline{54.18} & \underline{46.44} & \textbf{73.40} & \underline{45.70} & \underline{64.79} & \underline{57.58} \\
      & BIOMEDICA$^{13.65\mathrm{B}}$ & 45.09 & 28.67 & \underline{59.00} & 79.09 & \textbf{56.36} & 53.64 & 44.67 & 71.72 & 42.90 & 63.48 & 55.69 \\
      & BIOMEDICA$^{9.63\mathrm{B}}$ & 44.06 & \underline{29.37} & 58.20 & 79.08 & 53.36 & 52.81 & 45.00 & 71.76 & 42.00 & 61.83 & 55.15 \\
      & PMC-InterCPT$^{9.63\mathrm{B}}$ & \textbf{47.20} & \textbf{32.17} & \textbf{59.30} & \textbf{81.93} & 55.23 & \textbf{55.17} & \textbf{46.78} & \underline{73.12} & \textbf{46.90} & \textbf{65.33} & \textbf{58.03} \\
      \bottomrule
    \end{tabular*}
  }
\end{table}

Against the token-matched raw control, PMC-InterCPT improves the medical average by 1.46 points and the general/scientific average by 3.11 points on Qwen3.5-4B-Base. The gains span distinct evaluation formats rather than a single benchmark protocol. The strongest matched-control gains are on MMMU-Med (+2.34), PretexEval (+3.42), ChartQA (+2.44), and CharXiv (+8.80). The full raw BIOMEDICA run does not close this gap: despite using 42\% more CPT tokens than PMC-InterCPT, it remains 1.58 points lower on the medical average and 2.44 points lower on the general/scientific average. Relative to Base+SFT, PMC-InterCPT raises the medical average by 1.56 points while maintaining general/scientific performance.

The same data comparison transfers to Qwen3.5-2B-Base. PMC-InterCPT improves the medical average by 2.35 points and the general/scientific average by 2.88 points over its token-matched raw control, and by 0.99 and 0.45 points over SFT-only. Together, the two scale settings indicate that source-grounded curation contributes beyond simply increasing the volume of in-domain raw data.

\subsection{Curation and Allocation Ablation}
\label{sec:mixture}

\begin{figure}[t]
  \centering
  \includegraphics[width=\textwidth]{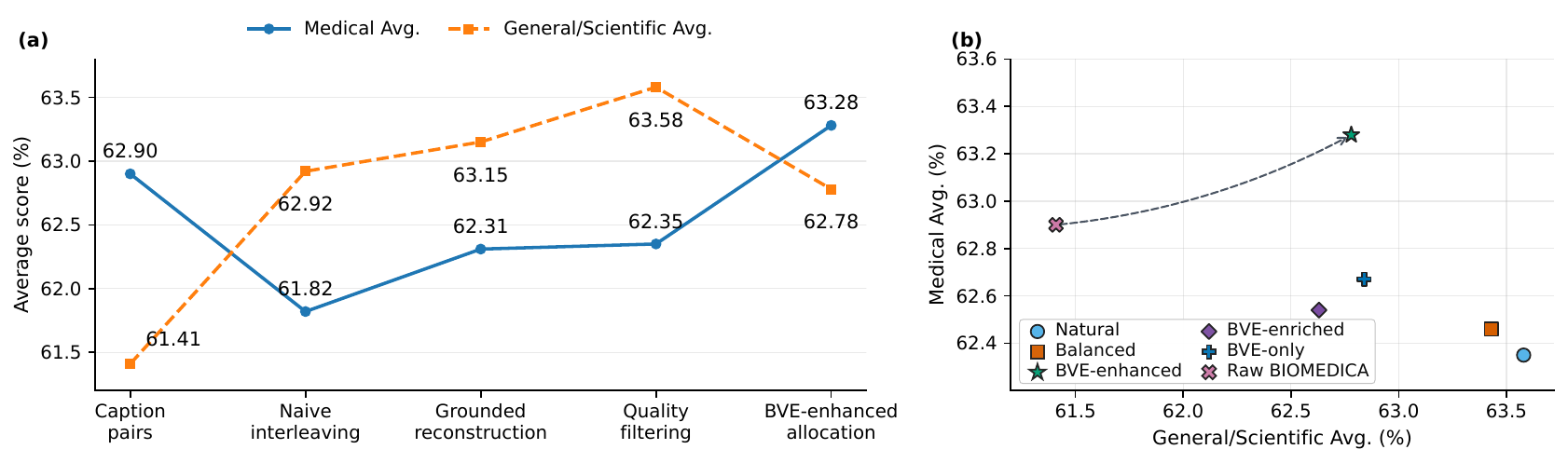}
  \caption{Component and mixture analysis with 1B tokens of CPT. (a) Medical and general/scientific averages as pipeline stages are added. (b) Medical versus general/scientific averages for the five mixtures and the Raw BIOMEDICA control. The dashed arrow points from Raw BIOMEDICA to BVE-enhanced.}
  \label{fig:component-mixture}
\end{figure}

Table~\ref{tab:component-ablation} reports benchmark-level results, while Figure~\ref{fig:component-mixture} shows the cumulative ablation and compares the medical and general/scientific averages across the five mixtures. The cumulative comparison uses the same backbone, 1B-token CPT budget, SFT data, and training hyperparameters while adding one pipeline stage at a time. This design distinguishes raw context addition from the subsequent repair, filtering, and evidence-allocation choices.

\begin{table}[t]
  \caption{Controlled pipeline ablation. Superscripts give CPT token budgets. Rows introduce interleaving, context reconstruction, and quality filtering; BVE-enh. denotes the BVE-enhanced mixture.}
  \label{tab:component-ablation}
  \vspace{3pt}
  \centering
  {\fontsize{6}{7}\selectfont
    \setlength{\tabcolsep}{0.65pt}
    \begin{tabular*}{\textwidth}{@{\extracolsep{\fill}}cccc*{11}{c}}
      \toprule
      \textbf{Format} & \textbf{Clean} & \textbf{QF} & \textbf{Mixture} & \multicolumn{6}{c}{\textbf{Medical benchmarks}} & \multicolumn{5}{c}{\textbf{General/scientific benchmarks}} \\
      \cmidrule(lr){5-10}\cmidrule(lr){11-15}
      & & & & \textbf{\shortstack{MMMU\\Med}} & \textbf{\shortstack{MMMU-Pro\\Med}} & \textbf{\shortstack{PMC\\VQA}} & \textbf{\shortstack{Omni\\Med}} & \textbf{\shortstack{Pretex\\Eval}} & \textbf{\shortstack{Medical\\avg.}} & \textbf{\shortstack{MMMU\\All}} & \textbf{\shortstack{Chart\\QA}} & \textbf{\shortstack{Char\\Xiv}} & \textbf{\shortstack{Sci\\VQA}} & \textbf{\shortstack{General\\avg.}} \\
      \midrule
      Caption$^{1\mathrm{B}}$ & -- & -- & -- & 58.39 & \textbf{42.66} & 60.90 & 86.04 & \underline{66.50} & \underline{62.90} & 54.78 & 75.88 & 48.20 & 66.78 & 61.41 \\
      Interleaved$^{1\mathrm{B}}$ & No & No & Natural & 59.30 & 37.41 & \underline{61.40} & 85.73 & 65.24 & 61.82 & 54.11 & 76.56 & 52.00 & \textbf{69.00} & 62.92 \\
      Interleaved$^{1\mathrm{B}}$ & Yes & No & Natural & \underline{59.53} & 39.86 & 61.20 & 86.17 & 64.77 & 62.31 & 55.22 & 76.12 & \underline{52.70} & 68.55 & \underline{63.15} \\
      Interleaved$^{1\mathrm{B}}$ & Yes & Yes & Natural & \textbf{60.10} & 41.26 & 61.10 & \textbf{86.38} & 62.92 & 62.35 & \underline{55.33} & \textbf{77.08} & \textbf{53.30} & \underline{68.60} & \textbf{63.58} \\
      Interleaved$^{1\mathrm{B}}$ & Yes & Yes & BVE-enh. & 59.36 & \underline{41.96} & \textbf{61.50} & \underline{86.19} & \textbf{67.37} & \textbf{63.28} & \textbf{55.44} & \underline{76.64} & 51.60 & 67.45 & 62.78 \\
      \bottomrule
    \end{tabular*}
  }
\end{table}

Naively interleaving article context improves the general/scientific average by 1.51 points but decreases the medical average by 1.08 points relative to raw caption pairs. Thus, article text alone is not a useful substitute for figure-grounded context. Context reconstruction recovers 0.49 points on the medical average and 0.23 points on the general/scientific average. Quality filtering then adds 0.05 points on the medical average and 0.43 points on the general/scientific average.

\paragraph{One-Billion-Token Mixture Configurations.}

Natural preserves the source-pool distribution (approximately 28\% BVE, 52\% QTE, 17\% MSE, and 3\% AUX). Balanced assigns 33\%, 33\%, 29\%, and 5\% to the four buckets. We set BVE-enhanced to 45/30/20/5 to make BVE the largest bucket and reserve 55\% of tokens for the other evidence types. BVE-enriched raises BVE to 70\% and splits the remaining budget at the same QTE:MSE:AUX ratio of 6:4:1, giving target shares of approximately 16.36\%, 10.91\%, and 2.73\%. BVE-only assigns the full budget to BVE.

\begin{table}[t]
  \caption{Comparison of five mixture configurations at a 1B-token CPT budget. The mixtures use the same clean, quality-filtered PMC-InterCPT source pool and fixed LLaVA-OneVision SFT stage. Raw BIOMEDICA is a caption-pair control.}
  \label{tab:mixture-configurations}
  \vspace{3pt}
  \centering
  {\fontsize{6}{7}\selectfont
    \setlength{\tabcolsep}{0.75pt}
    \begin{tabular*}{\textwidth}{@{\extracolsep{\fill}}cc*{11}{c}}
      \toprule
      \textbf{Backbone} & \textbf{CPT data} & \multicolumn{6}{c}{\textbf{Medical benchmarks}} & \multicolumn{5}{c}{\textbf{General/scientific benchmarks}} \\
      \cmidrule(lr){3-8}\cmidrule(lr){9-13}
      & & \textbf{\shortstack{MMMU\\Med}} & \textbf{\shortstack{MMMU-Pro\\Med}} & \textbf{\shortstack{PMC\\VQA}} & \textbf{\shortstack{Omni\\Med}} & \textbf{\shortstack{Pretex\\Eval}} & \textbf{\shortstack{Medical\\avg.}} & \textbf{\shortstack{MMMU\\All}} & \textbf{\shortstack{Chart\\QA}} & \textbf{\shortstack{Char\\Xiv}} & \textbf{\shortstack{Sci\\VQA}} & \textbf{\shortstack{General\\avg.}} \\
      \midrule
      \multirow{6}{*}{\shortstack{Qwen3.5-\\4B-Base}} & Raw BIOMEDICA$^{1\mathrm{B}}$ & 58.39 & \textbf{42.66} & 60.90 & 86.04 & \underline{66.50} & \underline{62.90} & 54.78 & 75.88 & 48.20 & 66.78 & 61.41 \\
      \cmidrule(lr){2-13}
      & Natural$^{1\mathrm{B}}$ & \textbf{60.10} & 41.26 & 61.10 & 86.38 & 62.92 & 62.35 & \underline{55.33} & 77.08 & \textbf{53.30} & \underline{68.60} & \textbf{63.58} \\
      & Balanced$^{1\mathrm{B}}$ & 59.42 & 39.86 & \textbf{62.80} & 85.56 & 64.65 & 62.46 & \textbf{55.44} & 76.92 & \underline{52.60} & \textbf{68.74} & \underline{63.43} \\
      & BVE-enhanced$^{1\mathrm{B}}$ & 59.36 & \underline{41.96} & 61.50 & 86.19 & \textbf{67.37} & \textbf{63.28} & \textbf{55.44} & 76.64 & 51.60 & 67.45 & 62.78 \\
      & BVE-enriched$^{1\mathrm{B}}$ & \underline{59.93} & 41.26 & 60.10 & \textbf{86.64} & 64.78 & 62.54 & 54.11 & \textbf{78.12} & 52.40 & 65.88 & 62.63 \\
      & BVE-only$^{1\mathrm{B}}$ & 59.82 & 39.51 & \underline{61.75} & \underline{86.51} & 65.76 & 62.67 & 54.78 & \underline{77.20} & 51.50 & 67.88 & 62.84 \\
      \bottomrule
    \end{tabular*}
  }
\end{table}

Table~\ref{tab:mixture-configurations} show that the medical average rises from 62.35 at Natural to 62.46 at 33\% and reaches 63.28 at 45\%. It falls to 62.54 at 70\%. Natural achieves the highest general/scientific average. Increasing BVE from 45\% to 70\% improves OmniMedVQA and ChartQA by 0.45 and 1.48 points, while PretexEval and PMC-VQA-clean fall by 2.59 and 1.40 points.

At 100\% BVE, OmniMedVQA and PMC-VQA-clean gain 0.32 and 0.25 points over the 45\% mixture, while MMMU-Pro Med and PretexEval lose 2.45 and 1.61 points. The general/scientific average changes by only 0.06 points (62.78 to 62.84), while the medical average falls by 0.61 points.

\begin{figure}[t]
  \centering
  \includegraphics[width=\textwidth]{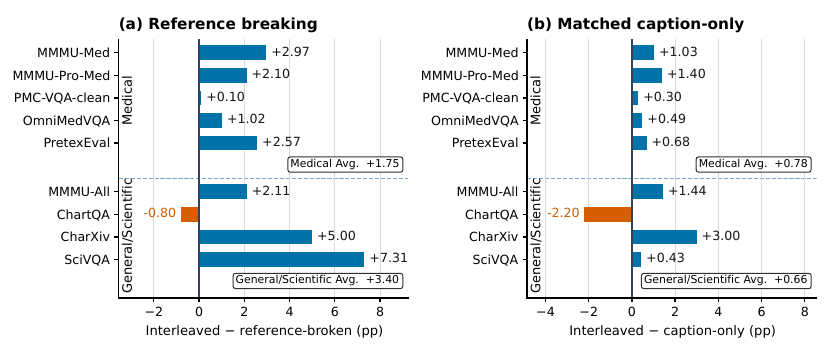}
  \caption{Figure-context controls for the 1B-token BVE-enhanced mixture. Bars show interleaved scores minus (a) reference-broken or (b) mixture-matched caption-only scores. Boxes give group averages. The panels use different horizontal scales.}
  \label{fig:reference-breaking}
\end{figure}

BVE accounts for 28.13\% of the filtered source pool, or 4.33B tokens (Table~\ref{tab:source-pool-buckets}). We avoid repeating records to reach a target share because repeated records consume CPT tokens without adding new evidence. The available BVE limits a 70\% mixture to roughly 6.2B tokens and a BVE-only mixture to 4.33B tokens. We chose the 45\% BVE-enhanced mixture because it achieved the highest medical average across the five 1B-token settings and supported a 9.63B-token corpus with 55\% of its budget reserved for quantitative, mechanistic, and auxiliary evidence.

\subsection{Validation of Figure-Context Grounding}
\label{sec:reference-breaking}

\begin{table}[t]
  \caption{Comparison of biomedical corpora with approximately 1B tokens of CPT on Qwen3.5-4B-Base. Scores are percentages. Bold and underlining mark the best and second-best values in each column.}
  \label{tab:public-corpus-comparison}
  \vspace{3pt}
  \centering
  {\fontsize{6}{7}\selectfont
    \setlength{\tabcolsep}{0.75pt}
    %BEGIN DIFNOMARKUP
    \begin{tabular*}{\textwidth}{@{\extracolsep{\fill}}l*{11}{c}}
      \toprule
      \textbf{CPT data} & \multicolumn{6}{c}{\textbf{Medical benchmarks}} & \multicolumn{5}{c}{\textbf{General/scientific benchmarks}} \\
      \cmidrule(lr){2-7}\cmidrule(lr){8-12}
      & \textbf{\shortstack{MMMU\\Med}} & \textbf{\shortstack{MMMU-Pro\\Med}} & \textbf{\shortstack{PMC\\VQA}} & \textbf{\shortstack{Omni\\Med}} & \textbf{\shortstack{Pretex\\Eval}} & \textbf{\shortstack{Medical\\avg.}} & \textbf{\shortstack{MMMU\\All}} & \textbf{\shortstack{Chart\\QA}} & \textbf{\shortstack{Char\\Xiv}} & \textbf{\shortstack{Sci\\VQA}} & \textbf{\shortstack{General\\avg.}} \\
      \midrule
      %END DIFNOMARKUP
      Raw BIOMEDICA & 58.39 & \textbf{42.66} & 60.90 & 86.04 & 66.50 & \underline{62.90} & 54.78 & 75.88 & 48.20 & 66.78 & 61.41 \\
      Open-PMC & 57.31 & 35.31 & 60.55 & 84.97 & 66.68 & 60.96 & 54.78 & \underline{77.28} & 49.80 & 67.14 & 62.25 \\
      Open-PMC-18M & 57.65 & 37.41 & \underline{61.40} & 85.50 & \textbf{70.09} & 62.41 & 53.89 & \textbf{78.60} & 49.60 & \underline{67.48} & 62.39 \\
      \midrule
      PMC-InterCPT (Natural) & \textbf{60.10} & 41.26 & 61.10 & \textbf{86.38} & 62.92 & 62.35 & \underline{55.33} & 77.08 & \textbf{53.30} & \textbf{68.60} & \textbf{63.58} \\
      PMC-InterCPT (BVE-enhanced) & \underline{59.36} & \underline{41.96} & \textbf{61.50} & \underline{86.19} & \underline{67.37} & \textbf{63.28} & \textbf{55.44} & 76.64 & \underline{51.60} & 67.45 & \underline{62.78} \\
      \bottomrule
    \end{tabular*}
  }
\end{table}

\paragraph{Reference-Breaking Control.}
We test whether the original figure--context links matter by replacing each eligible context segment with a random text span from the same article with a matching token count. The control preserves the images, captions, source articles, and token budget. Figure~\ref{fig:reference-breaking}(a) shows that breaking these links lowers the medical and general/scientific averages by 1.75 and 3.40 points, respectively. SciVQA and CharXiv show the largest drops, at 7.31 and 5.00 points, respectively. Scores decline on all five medical benchmarks.

\paragraph{Mixture-Matched Caption-Only Control.}
The reference-breaking control tests whether downstream performance depends on pairing figures with their original body-text context. The caption-only control tests whether adding figure-linked body text improves performance over images and captions alone at a matched evidence allocation and CPT token budget. We draw the caption-only and interleaved mixtures from the same filtered source pool. We retain images and captions, remove the body-text context, and recompute token counts before resampling to the same 45/30/20/5 BVE/QTE/MSE/AUX token allocation and approximately 1B CPT tokens. Both runs use Qwen3.5-4B-Base and the same SFT setting.

Figure~\ref{fig:reference-breaking}(b) shows that the interleaved run improves all five medical benchmarks. The gains are 1.03 points on MMMU-Med, 1.40 on MMMU-Pro-Med, 0.30 on PMC-VQA-clean, 0.49 on OmniMedVQA, and 0.68 on PretexEval, for an average gain of 0.78 points. The general/scientific average rises by 0.66 points, from 62.12 to 62.78. CharXiv gains 3.00 points, MMMU-All gains 1.44, and SciVQA gains 0.43, while ChartQA falls by 2.20 points. The gains across all five medical benchmarks in both comparisons support the use of reconstructed, figure-linked context in continued pretraining of medical multimodal foundation models. Full results are in the appendix.

\subsection{Comparison with Public Biomedical Corpora}
\label{sec:public-corpus-comparison}

We compare the Natural and BVE-enhanced allocations of PMC-InterCPT with Open-PMC~\citep{baghbanzadeh2025advancing} and Open-PMC-18M~\citep{baghbanzadeh2025open}, two public biomedical corpora that enrich image--text pairs using article context. We sample about 1B multimodal tokens each from Open-PMC (context-enriched text) and Open-PMC-18M (subfigure--text pairs). All rows in Table~\ref{tab:public-corpus-comparison} use Qwen3.5-4B-Base, the same SFT data, and approximately 1B multimodal CPT tokens. Raw BIOMEDICA provides a caption-pair baseline.

The BVE-enhanced mixture achieves the highest medical average (63.28), ahead of Open-PMC, Open-PMC-18M, and raw BIOMEDICA by 2.32, 0.87, and 0.38 points, respectively. PMC-InterCPT Natural is close to Open-PMC-18M on the medical average (62.35 versus 62.41) and scores 1.19 points higher on the general/scientific average. Among the five settings, Natural has the highest general/scientific average, and BVE-enhanced has the highest medical average, 0.38 points above raw BIOMEDICA. These results highlight the value of PMC-InterCPT as a biomedical pretraining corpus for both medical and general/scientific multimodal tasks.

\subsection{Cross-Architecture Ablation}
\label{sec:transfer}

Table~\ref{tab:llava-transfer} tests whether the corpus-construction benefit extends to LLaVA-OneVision-1.5-4B-Base under a fixed 1B-token CPT budget. Relative to the token-matched raw BIOMEDICA control, PMC-InterCPT improves the medical average by 1.63 points and the general/scientific average by 1.37 points. The improvement covers four of the five medical benchmarks and is especially pronounced on SciVQA (+6.57), suggesting that the benefit is not confined to the Qwen architecture.

\begin{table}[t]
  \caption{Cross-architecture ablation on LLaVA-OneVision-1.5-4B-Base. Superscripts give CPT token budgets; all CPT rows use the same SFT stage.}
  \label{tab:llava-transfer}
  \vspace{3pt}
  \centering
  {\fontsize{6}{7}\selectfont
    \setlength{\tabcolsep}{0.65pt}
    \begin{tabular*}{\textwidth}{@{\extracolsep{\fill}}cc*{11}{c}}
      \toprule
      \textbf{Backbone} & \textbf{CPT data} & \multicolumn{6}{c}{\textbf{Medical benchmarks}} & \multicolumn{5}{c}{\textbf{General/scientific benchmarks}} \\
      \cmidrule(lr){3-8}\cmidrule(lr){9-13}
      & & \textbf{\shortstack{MMMU\\Med}} & \textbf{\shortstack{MMMU-Pro\\Med}} & \textbf{\shortstack{PMC\\VQA}} & \textbf{\shortstack{Omni\\Med}} & \textbf{\shortstack{Pretex\\Eval}} & \textbf{\shortstack{Medical\\avg.}} & \textbf{\shortstack{MMMU\\All}} & \textbf{\shortstack{Chart\\QA}} & \textbf{\shortstack{Char\\Xiv}} & \textbf{\shortstack{Sci\\VQA}} & \textbf{\shortstack{General\\avg.}} \\
      \midrule
      \multirow{3}{*}{\shortstack{LLaVA-OneVision\\1.5-4B-Base}} & None (SFT only) & 45.95 & \underline{25.87} & 48.45 & 62.44 & \underline{56.62} & 47.87 & 47.11 & \textbf{59.28} & \textbf{35.70} & \textbf{53.95} & \textbf{49.01} \\
      & BIOMEDICA$^{1\mathrm{B}}$ & \textbf{48.69} & 24.13 & \underline{50.25} & \underline{64.21} & 55.75 & \underline{48.61} & \underline{48.11} & \underline{58.24} & \underline{34.90} & 46.07 & 46.83 \\
      & PMC-InterCPT$^{1\mathrm{B}}$ & \underline{48.29} & \textbf{27.62} & \textbf{51.85} & \textbf{65.93} & \textbf{57.49} & \textbf{50.24} & \textbf{48.44} & 57.72 & 34.00 & \underline{52.64} & \underline{48.20} \\
      \bottomrule
    \end{tabular*}
  }
\end{table}

\begin{table}[t]
  \caption{Multi-image evaluation on Qwen3.5-4B-Base with fixed SFT. Superscripts give CPT token budgets. Scores are percentages.}
  \label{tab:multi-image-validation}
  \vspace{3pt}
  \centering
  \small
  \begin{tabular*}{\linewidth}{@{\extracolsep{\fill}}lccccc@{}}
    \toprule
    \textbf{CPT data} & \textbf{MedFrameQA} & \textbf{MedThinkVQA} & \textbf{MuirBench} & \textbf{MultiChartQA} & \textbf{Overall} \\
    \midrule
    None (SFT only) & 57.63 & \underline{41.11} & 56.96 & 45.83 & 50.38 \\
    BIOMEDICA$^{13.65\mathrm{B}}$ & \underline{58.75} & 40.55 & 57.15 & 43.95 & 50.10 \\
    BIOMEDICA$^{9.63\mathrm{B}}$ & 58.61 & 40.28 & \underline{57.19} & \textbf{46.55} & \underline{50.66} \\
    PMC-InterCPT$^{9.63\mathrm{B}}$ & \textbf{59.35} & \textbf{43.06} & \textbf{58.27} & \underline{46.06} & \textbf{51.69} \\
    \bottomrule
  \end{tabular*}
\end{table}

\subsection{Multi-Image Evaluation}
\label{sec:multi-image-evaluation}

We evaluate the Qwen3.5-4B-Base checkpoints from Table~\ref{tab:main-results} on four benchmarks. MedFrameQA~\citep{yu2026medframeqa} and MedThinkVQA~\citep{yao2026medthinkvqa} test medical reasoning across images. MuirBench~\citep{wang2025muirbench} tests general multi-image understanding, and MultiChartQA~\citep{zhu2025multichartqa} tests reasoning across charts.

At the same 9.63B-token CPT budget, PMC-InterCPT improves the overall score by 1.03 percentage points over raw BIOMEDICA (Table~\ref{tab:multi-image-validation}). It gains 0.74 points on MedFrameQA, 2.78 on MedThinkVQA, and 1.08 on MuirBench, while the raw baseline leads by 0.49 points on MultiChartQA. PMC-InterCPT also improves over SFT alone on all four benchmarks. These results show that the gains from PMC-InterCPT extend to medical and general tasks involving multiple images.

\section{Conclusion}
We introduced PMC-InterCPT, a sequential framework for constructing biomedical interleaved CPT data. It uses source figure references to reconstruct and repair coherent sequences, filters the reconstructed records for textual usability and medical relevance, and then allocates the retained data through a controllable evidence mixture. Its 9.63B-token instantiation improves over the raw source pool on medical and general multimodal performance under a fixed CPT+SFT protocol. The component ablation further shows that article context becomes useful only after referent-aware reconstruction and curation. We hope this framework supports more reliable use of scientific literature.

\subsection*{AI use statement}
Generative AI tools were used for brainstorming, information gathering, code development assistance, improving manuscript readability, identifying relevant literature, and formatting references. A generative model was also used as the annotation model described in the medical-relevance and text-quality classification stage. The authors reviewed and verified all AI-assisted code and annotations, and take responsibility for the final content and claims of this work.

\subsection*{Ethics statement}
PMC-InterCPT is derived from scientific articles in PMC-OA and does not involve new data collection from human participants. Corpus construction preserves source-level license metadata, and any public redistribution should be restricted to records whose licenses permit it. Biomedical publications may contain sensitive clinical descriptions and demographic or publication biases; PMC-InterCPT is intended for research, not unvalidated clinical use.

\subsection*{Reproducibility statement}
The main paper specifies the corpus-construction stages, controlled data comparisons, training protocol, and evaluation suite. The appendix provides detailed dataset statistics, reconstruction and filtering rules, annotation prompts, classifier settings, benchmark protocols, and complete auxiliary results. We also provide a code and data supplement containing the processing pipeline and representative reconstructed samples. If accepted, we will publicly release PMC-InterCPT in accordance with source-license and redistribution requirements.

\subsection*{Acknowledgments}
This work was supported by the Hong Kong RGC (TRS: T41-517/25-N; GRF: 15228325), ITC RAISe+ (RAI/24/1/086A), and the Research Institute for Generative Artificial Intelligence at PolyU.

\bibliography{pmc_intercpt}

\begin{thebibliography}{36}
\providecommand{\natexlab}[1]{#1}
\providecommand{\url}[1]{\texttt{#1}}
\expandafter\ifx\csname urlstyle\endcsname\relax
  \providecommand{\doi}[1]{doi: #1}\else
  \providecommand{\doi}{doi: \begingroup \urlstyle{rm}\Url}\fi

\bibitem[Alayrac et~al.(2022)Alayrac, Donahue, Luc, Miech, Barr, Hasson, Lenc,
  Mensch, Millican, Reynolds, et~al.]{alayrac2022flamingo}
Jean-Baptiste Alayrac, Jeff Donahue, Pauline Luc, Antoine Miech, Iain Barr,
  Yana Hasson, Karel Lenc, Arthur Mensch, Katherine Millican, Malcolm Reynolds,
  et~al.
\newblock Flamingo: a visual language model for few-shot learning.
\newblock \emph{Advances in Neural Information Processing Systems},
  35:\penalty0 23716--23736, 2022.

\bibitem[Baghbanzadeh et~al.(2025{\natexlab{a}})Baghbanzadeh, Fallahpour,
  Parhizkar, Ogidi, Roy, Ashkezari, Khazaie, Colacci, Etemad, Afkanpour, and
  Dolatabadi]{baghbanzadeh2025advancing}
Negin Baghbanzadeh, Adibvafa Fallahpour, Yasaman Parhizkar, Franklin Ogidi,
  Shuvendu Roy, Sajad Ashkezari, Vahid~Reza Khazaie, Michael Colacci, Ali
  Etemad, Arash Afkanpour, and Elham Dolatabadi.
\newblock Advancing medical representation learning through high-quality data.
\newblock \emph{arXiv preprint arXiv:2503.14377}, 2025{\natexlab{a}}.

\bibitem[Baghbanzadeh et~al.(2025{\natexlab{b}})Baghbanzadeh, Islam, Ashkezari,
  Dolatabadi, and Afkanpour]{baghbanzadeh2025open}
Negin Baghbanzadeh, Mohammed~Saidul Islam, Sajad Ashkezari, Elham Dolatabadi,
  and Arash Afkanpour.
\newblock Open-pmc-18m: A high-fidelity large scale medical dataset for
  multimodal representation learning.
\newblock \emph{arXiv preprint arXiv:2506.02738}, 2025{\natexlab{b}}.

\bibitem[Bannur et~al.(2024)Bannur, Bouzid, Castro, Schwaighofer, Thieme,
  Bond-Taylor, Ilse, P{\'e}rez-Garc{\'i}a, Salvatelli, Sharma,
  et~al.]{bannur2024maira}
Shruthi Bannur, Kenza Bouzid, Daniel~C Castro, Anton Schwaighofer, Anja Thieme,
  Sam Bond-Taylor, Maximilian Ilse, Fernando P{\'e}rez-Garc{\'i}a, Valentina
  Salvatelli, Harshita Sharma, et~al.
\newblock {MAIRA-2}: Grounded radiology report generation.
\newblock \emph{arXiv preprint arXiv:2406.04449}, 2024.

\bibitem[Borisova et~al.(2025)Borisova, Rauscher, and Rehm]{borisova2025scivqa}
Ekaterina Borisova, Nikolas Rauscher, and Georg Rehm.
\newblock Scivqa 2025: Overview of the first scientific visual question
  answering shared task.
\newblock In \emph{Proceedings of the Fifth Workshop on Scholarly Document
  Processing (SDP 2025)}, pp.\  182--210, 2025.

\bibitem[Chen et~al.(2024)Chen, Gui, Ouyang, Gao, Chen, Chen, Wang, Zhang, Cai,
  Ji, Yu, Wan, and Wang]{chen2024huatuogptvision}
Junying Chen, Chi Gui, Ruyi Ouyang, Anningzhe Gao, Shunian Chen, Guiming~Hardy
  Chen, Xidong Wang, Ruifei Zhang, Zhenyang Cai, Ke~Ji, Guangjun Yu, Xiang Wan,
  and Benyou Wang.
\newblock {HuatuoGPT-Vision}, towards injecting medical visual knowledge into
  multimodal llms at scale.
\newblock \emph{arXiv preprint arXiv:2406.19280}, 2024.

\bibitem[Chen et~al.(2015)Chen, Zheng, Zhu, Shi, Tian, Zhao, Liu, H{\"u}ppi,
  {Troy II}, and Wang]{chen2015lactoferrin}
Yue Chen, Zhiqiang Zheng, Xi~Zhu, Yujie Shi, Dandan Tian, Fengjuan Zhao,
  Ni~Liu, Petra~S. H{\"u}ppi, Frederic~A. {Troy II}, and Bing Wang.
\newblock Lactoferrin promotes early neurodevelopment and cognition in
  postnatal piglets by upregulating the {BDNF} signaling pathway and
  polysialylation.
\newblock \emph{Molecular Neurobiology}, 52\penalty0 (1):\penalty0 256--269,
  2015.
\newblock \doi{10.1007/s12035-014-8856-9}.

\bibitem[Deng et~al.(2018)Deng, Wang, Long, Liu, Xiang, Bao, and
  Wang]{deng2018sirtuin}
Zhigang Deng, Xingbiao Wang, Xuan Long, Wanzhong Liu, Chunhua Xiang, Feng Bao,
  and Dong Wang.
\newblock Sirtuin 7 promotes colorectal carcinoma proliferation and invasion
  through the inhibition of {E-cadherin}.
\newblock \emph{Experimental and Therapeutic Medicine}, 15\penalty0
  (3):\penalty0 2333--2342, 2018.
\newblock \doi{10.3892/etm.2017.5673}.

\bibitem[Hu et~al.(2024)Hu, Li, Lu, Shao, He, Qiao, and Luo]{hu2024omnimedvqa}
Yutao Hu, Tianbin Li, Quanfeng Lu, Wenqi Shao, Junjun He, Yu~Qiao, and Ping
  Luo.
\newblock {OmniMedVQA}: A new large-scale comprehensive evaluation benchmark
  for medical {LVLM}.
\newblock \emph{Proceedings of the IEEE/CVF Conference on Computer Vision and
  Pattern Recognition}, pp.\  22170--22183, 2024.

\bibitem[Lauren{\c{c}}on et~al.(2023)Lauren{\c{c}}on, Saulnier, Tronchon,
  Bekman, Singh, Lozhkov, Wang, Karamcheti, Rush, Kiela,
  et~al.]{laurenccon2023obelics}
Hugo Lauren{\c{c}}on, Lucile Saulnier, L{\'e}o Tronchon, Stas Bekman, Amanpreet
  Singh, Anton Lozhkov, Thomas Wang, Siddharth Karamcheti, Alexander Rush,
  Douwe Kiela, et~al.
\newblock {OBELICS}: An open web-scale filtered dataset of interleaved
  image-text documents.
\newblock \emph{Advances in Neural Information Processing Systems},
  36:\penalty0 71683--71702, 2023.

\bibitem[Li et~al.(2024{\natexlab{a}})Li, Zhang, Guo, Zhang, Li, Zhang, Zhang,
  Zhang, Li, Liu, et~al.]{li2024llavaonevision}
Bo~Li, Yuanhan Zhang, Dong Guo, Renrui Zhang, Feng Li, Hao Zhang, Kaichen
  Zhang, Peiyuan Zhang, Yanwei Li, Ziwei Liu, et~al.
\newblock Llava-onevision: Easy visual task transfer.
\newblock \emph{arXiv preprint arXiv:2408.03326}, 2024{\natexlab{a}}.

\bibitem[Li et~al.(2023)Li, Wong, Zhang, Usuyama, Liu, Yang, Naumann, Poon, and
  Gao]{li2023llava}
Chunyuan Li, Cliff Wong, Sheng Zhang, Naoto Usuyama, Haotian Liu, Jianwei Yang,
  Tristan Naumann, Hoifung Poon, and Jianfeng Gao.
\newblock {LLaVA-Med}: Training a large language-and-vision assistant for
  biomedicine in one day.
\newblock \emph{Advances in Neural Information Processing Systems},
  36:\penalty0 28541--28564, 2023.

\bibitem[Li et~al.(2024{\natexlab{b}})Li, Fang, Smyrnis, Ivgi, Jordan, Gadre,
  Bansal, Guha, Keh, Arora, et~al.]{li2024datacomp}
Jeffrey Li, Alex Fang, Georgios Smyrnis, Maor Ivgi, Matt Jordan, Samir Gadre,
  Hritik Bansal, Etash Guha, Sedrick Keh, Kushal Arora, et~al.
\newblock {DataComp-LM}: In search of the next generation of training sets for
  language models.
\newblock \emph{Advances in Neural Information Processing Systems},
  37:\penalty0 14200--14282, 2024{\natexlab{b}}.

\bibitem[Lin et~al.(2024)Lin, Yin, Ping, Molchanov, Shoeybi, and
  Han]{lin2024vila}
Ji~Lin, Hongxu Yin, Wei Ping, Pavlo Molchanov, Mohammad Shoeybi, and Song Han.
\newblock Vila: On pre-training for visual language models.
\newblock In \emph{Proceedings of the IEEE/CVF Conference on Computer Vision
  and Pattern Recognition}, pp.\  26689--26699, 2024.

\bibitem[Lin et~al.(2023)Lin, Zhao, Zhang, Wu, Zhang, Wang, and
  Xie]{lin2023pmcclip}
Weixiong Lin, Ziheng Zhao, Xiaoman Zhang, Chaoyi Wu, Ya~Zhang, Yanfeng Wang,
  and Weidi Xie.
\newblock Pmc-clip: Contrastive language-image pre-training using biomedical
  documents.
\newblock In \emph{International Conference on Medical Image Computing and
  Computer-Assisted Intervention}, pp.\  525--536. Springer, 2023.

\bibitem[Lozano et~al.(2025)Lozano, Sun, Burgess, Chen, Nirschl, Gu, Lopez,
  Aklilu, Rau, Katzer, et~al.]{lozano2025biomedica}
Alejandro Lozano, Min~Woo Sun, James Burgess, Liangyu Chen, Jeffrey~J Nirschl,
  Jeffrey Gu, Ivan Lopez, Josiah Aklilu, Anita Rau, Austin~Wolfgang Katzer,
  et~al.
\newblock Biomedica: An open biomedical image-caption archive, dataset, and
  vision-language models derived from scientific literature.
\newblock In \emph{Proceedings of the Computer Vision and Pattern Recognition
  Conference}, pp.\  19724--19735, 2025.

\bibitem[Masry et~al.(2022)Masry, Long, Tan, Joty, and Hoque]{masry2022chartqa}
Ahmed Masry, Do~Long, Jia~Qing Tan, Shafiq Joty, and Enamul Hoque.
\newblock {ChartQA}: A benchmark for question answering about charts with
  visual and logical reasoning.
\newblock In \emph{Findings of the Association for Computational Linguistics:
  ACL 2022}, pp.\  2263--2279, 2022.

\bibitem[Moor et~al.(2023)Moor, Huang, Wu, Yasunaga, Dalmia, Leskovec, Zakka,
  Reis, and Rajpurkar]{moor2023med}
Michael Moor, Qian Huang, Shirley Wu, Michihiro Yasunaga, Yash Dalmia, Jure
  Leskovec, Cyril Zakka, Eduardo~Pontes Reis, and Pranav Rajpurkar.
\newblock {Med-Flamingo}: a multimodal medical few-shot learner.
\newblock In \emph{Machine Learning for Health (ML4H)}, pp.\  353--367. PMLR,
  2023.

\bibitem[Penedo et~al.(2023)Penedo, Malartic, Hesslow, Cojocaru, Cappelli,
  Alobeidli, Pannier, Almazrouei, and Launay]{penedo2023refinedweb}
Guilherme Penedo, Quentin Malartic, Daniel Hesslow, Ruxandra Cojocaru,
  Alessandro Cappelli, Hamza Alobeidli, Baptiste Pannier, Ebtesam Almazrouei,
  and Julien Launay.
\newblock The {RefinedWeb} dataset for falcon {LLM}: Outperforming curated
  corpora with web data, and web data only.
\newblock \emph{arXiv preprint arXiv:2306.01116}, 2023.

\bibitem[Penedo et~al.(2024)Penedo, Kydl{\'i}{\v{c}}ek, Lozhkov, Mitchell,
  Raffel, Von~Werra, Wolf, et~al.]{penedo2024fineweb}
Guilherme Penedo, Hynek Kydl{\'i}{\v{c}}ek, Anton Lozhkov, Margaret Mitchell,
  Colin Raffel, Leandro Von~Werra, Thomas Wolf, et~al.
\newblock The {FineWeb} datasets: Decanting the web for the finest text data at
  scale.
\newblock \emph{Advances in Neural Information Processing Systems},
  37:\penalty0 30811--30849, 2024.

\bibitem[{Qwen Team}(2026)]{qwen2026qwen35}
{Qwen Team}.
\newblock {Qwen3.5}: Towards native multimodal agents.
\newblock Alibaba Cloud Community Blog, 2026.
\newblock URL
  \url{https://www.alibabacloud.com/blog/qwen3-5-towards-native-multimodal-agents_602894}.

\bibitem[Wang et~al.(2025)Wang, Fu, Huang, Li, Liu, Liu, Ma, Xu, Zhou, Zhang,
  Yan, Mo, Liu, Lu, Li, Xiao, Chang, Roth, Zhang, Poon, and
  Chen]{wang2025muirbench}
Fei Wang, Xingyu Fu, James~Y. Huang, Zekun Li, Qin Liu, Xiaogeng Liu,
  Mingyu~Derek Ma, Nan Xu, Wenxuan Zhou, Kai Zhang, Tianyi~Lorena Yan,
  Wenjie~Jacky Mo, Hsiang-Hui Liu, Pan Lu, Chunyuan Li, Chaowei Xiao, Kai-Wei
  Chang, Dan Roth, Sheng Zhang, Hoifung Poon, and Muhao Chen.
\newblock {MuirBench}: A comprehensive benchmark for robust multi-image
  understanding.
\newblock In \emph{International Conference on Learning Representations}, 2025.

\bibitem[Wang et~al.(2024)Wang, Xia, He, Chen, Liu, Zhu, Liang, Wu, Liu,
  Malladi, et~al.]{wang2024charxiv}
Zirui Wang, Mengzhou Xia, Luxi He, Howard Chen, Yitao Liu, Richard Zhu, Kaiqu
  Liang, Xindi Wu, Haotian Liu, Sadhika Malladi, et~al.
\newblock Charxiv: Charting gaps in realistic chart understanding in multimodal
  llms.
\newblock \emph{Advances in Neural Information Processing Systems},
  37:\penalty0 113569--113697, 2024.

\bibitem[Xie et~al.(2023)Xie, Pham, Dong, Du, Liu, Lu, Liang, Le, Ma, and
  Yu]{xie2023doremi}
Sang~Michael Xie, Hieu Pham, Xuanyi Dong, Nan Du, Hanxiao Liu, Yifeng Lu,
  Percy~S Liang, Quoc~V Le, Tengyu Ma, and Adams~Wei Yu.
\newblock {DoReMi}: Optimizing data mixtures speeds up language model
  pretraining.
\newblock \emph{Advances in Neural Information Processing Systems},
  36:\penalty0 69798--69818, 2023.

\bibitem[Xu et~al.(2025)Xu, Chan, Li, Aljunied, Yuan, Wang, Xiao, Chen, Liu,
  Li, et~al.]{xu2025lingshu}
Weiwen Xu, Hou~Pong Chan, Long Li, Mahani Aljunied, Ruifeng Yuan, Jianyu Wang,
  Chenghao Xiao, Guizhen Chen, Chaoqun Liu, Zhaodonghui Li, et~al.
\newblock {Lingshu}: A generalist foundation model for unified multimodal
  medical understanding and reasoning.
\newblock \emph{arXiv preprint arXiv:2506.07044}, 2025.

\bibitem[Yao et~al.(2026)Yao, Wang, Zhang, Wang, Xia, Tang, Han, Ouyang, Yang,
  Cohan, and Yu]{yao2026medthinkvqa}
Zonghai Yao, Benlu Wang, Yifan Zhang, Junda Wang, Iris Xia, Zhipeng Tang, Shuo
  Han, Feiyun Ouyang, Zhichao Yang, Arman Cohan, and Hong Yu.
\newblock Medical thinking with multiple images.
\newblock \emph{arXiv preprint arXiv:2604.16506}, 2026.

\bibitem[Yu et~al.(2026)Yu, Wang, Wu, Luo, Wang, Xie, Rajpurkar, Yang, Yang,
  Wang, Yu, and Zhou]{yu2026medframeqa}
Suhao Yu, Haojin Wang, Juncheng Wu, Luyang Luo, Jingshen Wang, Cihang Xie,
  Pranav Rajpurkar, Carl Yang, Yang Yang, Kang Wang, Yannan Yu, and Yuyin Zhou.
\newblock {MedFrameQA}: A multi-image medical vqa benchmark for clinical
  reasoning.
\newblock In \emph{Proceedings of the 2026 Conference on Empirical Methods in
  Natural Language Processing}, 2026.

\bibitem[Yue et~al.(2024)Yue, Ni, Zhang, Zheng, Liu, Zhang, Stevens, Jiang,
  Ren, Sun, et~al.]{yue2024mmmu}
Xiang Yue, Yuansheng Ni, Kai Zhang, Tianyu Zheng, Ruoqi Liu, Ge~Zhang, Samuel
  Stevens, Dongfu Jiang, Weiming Ren, Yuxuan Sun, et~al.
\newblock {MMMU}: A massive multi-discipline multimodal understanding and
  reasoning benchmark for expert agi.
\newblock \emph{Proceedings of the IEEE/CVF Conference on Computer Vision and
  Pattern Recognition}, pp.\  9556--9567, 2024.

\bibitem[Yue et~al.(2025)Yue, Zheng, Ni, Wang, Zhang, Tong, Sun, Yu, Zhang,
  Sun, et~al.]{yue2025mmmu}
Xiang Yue, Tianyu Zheng, Yuansheng Ni, Yubo Wang, Kai Zhang, Shengbang Tong,
  Yuxuan Sun, Botao Yu, Ge~Zhang, Huan Sun, et~al.
\newblock Mmmu-pro: A more robust multi-discipline multimodal understanding
  benchmark.
\newblock In \emph{Proceedings of the 63rd Annual Meeting of the Association
  for Computational Linguistics (Volume 1: Long Papers)}, pp.\  15134--15186,
  2025.

\bibitem[Zhang et~al.(2023{\natexlab{a}})Zhang, Xu, Usuyama, Xu, Bagga, Tinn,
  Preston, Rao, Wei, Valluri, et~al.]{zhang2023biomedclip}
Sheng Zhang, Yanbo Xu, Naoto Usuyama, Hanwen Xu, Jaspreet Bagga, Robert Tinn,
  Sam Preston, Rajesh Rao, Mu~Wei, Naveen Valluri, et~al.
\newblock {BiomedCLIP}: a multimodal biomedical foundation model pretrained
  from fifteen million scientific image-text pairs.
\newblock \emph{arXiv preprint arXiv:2303.00915}, 2023{\natexlab{a}}.

\bibitem[Zhang et~al.(2023{\natexlab{b}})Zhang, Wu, Zhao, Lin, Zhang, Wang, and
  Xie]{zhang2023pmcvqa}
Xiaoman Zhang, Chaoyi Wu, Ziheng Zhao, Weixiong Lin, Ya~Zhang, Yanfeng Wang,
  and Weidi Xie.
\newblock {PMC-VQA}: Visual instruction tuning for medical visual question
  answering.
\newblock \emph{arXiv preprint arXiv:2305.10415}, 2023{\natexlab{b}}.

\bibitem[Zheng et~al.(2024{\natexlab{a}})Zheng, Yin, Xie, Sun, Huang, Yu, Cao,
  Kozyrakis, Stoica, Gonzalez, et~al.]{zheng2024sglang}
Lianmin Zheng, Liangsheng Yin, Zhiqiang Xie, Chuyue Sun, Jeff Huang, Cody~H Yu,
  Shiyi Cao, Christos Kozyrakis, Ion Stoica, Joseph~E Gonzalez, et~al.
\newblock Sglang: Efficient execution of structured language model programs.
\newblock \emph{Advances in neural information processing systems},
  37:\penalty0 62557--62583, 2024{\natexlab{a}}.

\bibitem[Zheng et~al.(2024{\natexlab{b}})Zheng, Zhang, Zhang, Ye, and
  Luo]{zheng2024llamafactory}
Yaowei Zheng, Richong Zhang, Junhao Zhang, Yanhan Ye, and Zheyan Luo.
\newblock Llamafactory: Unified efficient fine-tuning of 100+ language models.
\newblock In \emph{Proceedings of the 62nd annual meeting of the association
  for computational linguistics (volume 3: system demonstrations)}, pp.\
  400--410, 2024{\natexlab{b}}.

\bibitem[Zhou et~al.(2025)Zhou, Liu, Ning, Zhang, and Wu]{zhou2024pretexeval}
Yuxuan Zhou, Xien Liu, Chen Ning, Xiao Zhang, and Ji~Wu.
\newblock Reliable and diverse evaluation of llm medical knowledge mastery.
\newblock In \emph{International Conference on Learning Representations},
  volume 2025, pp.\  57899--57922, 2025.

\bibitem[Zhu et~al.(2023)Zhu, Hessel, Awadalla, Gadre, Dodge, Fang, Yu,
  Schmidt, Wang, and Choi]{zhu2023multimodal}
Wanrong Zhu, Jack Hessel, Anas Awadalla, Samir~Yitzhak Gadre, Jesse Dodge, Alex
  Fang, Youngjae Yu, Ludwig Schmidt, William~Yang Wang, and Yejin Choi.
\newblock Multimodal {C4}: An open, billion-scale corpus of images interleaved
  with text.
\newblock \emph{Advances in Neural Information Processing Systems},
  36:\penalty0 8958--8974, 2023.

\bibitem[Zhu et~al.(2025)Zhu, Jia, Zhang, Li, and Jiang]{zhu2025multichartqa}
Zifeng Zhu, Mengzhao Jia, Zhihan Zhang, Lang Li, and Meng Jiang.
\newblock {MultiChartQA}: Benchmarking vision-language models on multi-chart
  problems.
\newblock In \emph{Proceedings of the 2025 Conference of the Nations of the
  Americas Chapter of the Association for Computational Linguistics}, pp.\
  11341--11359, 2025.

\end{thebibliography}
\bibliographystyle{iclr2027_conference}

\clearpage
\appendix

\section{Additional Dataset Statistics}
\label{app:dataset-statistics}

\subsection{Raw BIOMEDICA Statistics}
BIOMEDICA is built from the PubMed Central Open Access subset and provides an article-level archive serialized into image-caption training records~\citep{lozano2025biomedica}.
The released corpus contains over 6M full-text articles, over 24M image-caption pairs, and over 30M figure references.
Each record includes article-level metadata and image-level annotations, including article identifiers, title, abstract, journal, license, keywords, MeSH terms, citation metadata, image identifiers, image hashes, panel labels, and global/local taxonomy labels.
Table~\ref{tab:raw-biomedica-stats} summarizes the raw BIOMEDICA properties that are most relevant to our pipeline.

\begin{table}[h]
  \caption{Raw BIOMEDICA statistics reported by \citet{lozano2025biomedica}.}
  \label{tab:raw-biomedica-stats}
  \centering
  {\small
    \begin{tabularx}{\columnwidth}{
        Z{0.55\columnwidth}
        Y
      }
      \toprule
      \textbf{Property}                 & \textbf{Value}          \\
      \midrule
      Full-text articles                & 6M+                     \\
      Image-caption pairs               & 24M+                    \\
      Figure references                 & 30M+                    \\
      Metadata fields                   & 27                      \\
      Primary format in BIOMEDICA       & Image-caption pairs     \\
      Additional signal in our pipeline & Figure-referencing text \\
      \bottomrule
    \end{tabularx}
  }
\end{table}

\subsection{Cleaned Dataset Statistics}
Table~\ref{tab:caption-recovery-stats} reports the missing-caption recovery statistics.
We processed 2,537 tar archives containing 72,151,269 files, including 24,050,423 JSON records.
Among these JSON records, 1,987,649 were modified during missing-caption recovery, corresponding to an 8.26\% modification rate.
For records whose caption field required recovery, 6,238,761 captions were successfully recovered and 3,639,453 were not recovered, giving a success rate of 63.2\%.
Because one JSON record can contain multiple image slots, the number of recovered captions can exceed the number of modified records.
No XML files were missing for the affected articles.
Most recovered captions correspond to table-like figure entries whose captions are available in the PMC XML.
Unrecovered cases are largely entries such as graphical abstracts, formulae, or other media objects for which the source XML does not provide a caption that can be extracted.

\begin{table}[h]
  \caption{Missing-caption recovery statistics. The success rate is computed as recovered captions divided by recovered plus unrecovered captions.}
  \label{tab:caption-recovery-stats}
  \centering
  {\small
    \begin{tabularx}{\columnwidth}{
        Z{0.55\columnwidth}
        Y
      }
      \toprule
      \textbf{Statistic}     & \textbf{Value} \\
      \midrule
      Processed tar archives & 2,537          \\
      Processed files        & 72,151,269     \\
      JSON records           & 24,050,423     \\
      JSON records modified  & 1,987,649      \\
      Modification rate      & 8.26\%         \\
      Captions recovered     & 6,238,761      \\
      Captions unrecovered   & 3,639,453      \\
      Recovery success rate  & 63.2\%         \\
      Missing XML files      & 0              \\
      \bottomrule
    \end{tabularx}
  }
\end{table}

Repeated-text cleanup modified 22,844,977 of 24,050,423 JSON records (94.99\%).
This high modification rate reflects the prevalence of repeated short fragments and duplicated local context spans introduced when PMC XML paragraphs with inline figure references are flattened into plain text.

Table~\ref{tab:dataset-funnel-stats} summarizes the major scale changes across the later pipeline stages.
Interleaved reconstruction reduces the number of training records relative to raw BIOMEDICA by reorganizing article-level figures into context-grounded samples, but increases the token count by adding body-text context.
Quality filtering, decontamination, source-pool selection, and final mixture sampling then progressively concentrate the corpus into the final PMC-InterCPT training set.

\begin{table}[h]
  \caption{Dataset scale across the main corpus construction stages.}
  \label{tab:dataset-funnel-stats}
  \centering
  {\small
    \begin{tabularx}{\columnwidth}{
        Y
        Z{0.15\columnwidth}
        Z{0.15\columnwidth}
      }
      \toprule
      \textbf{Stage}                   & \textbf{Samples} & \textbf{Tokens} \\
      \midrule
      Raw BIOMEDICA                    & 24.04M           & 13.65B          \\
      After interleaved reconstruction & 20.16M           & 19.89B          \\
      After minimum-length filtering   & 20.02M           & 19.80B          \\
      Quality score $\geq$ 1           & 19.45M           & 18.78B          \\
      After decontamination            & 19.44M           & 18.68B          \\
      Source-pool construction         & 16.45M           & 15.41B          \\
      PMC-InterCPT final mixture       & 10.11M           & 9.63B           \\
      \bottomrule
    \end{tabularx}
  }
\end{table}

\subsection{Evidence Bucket Distribution}
Table~\ref{tab:source-pool-buckets} reports the evidence-bucket distribution after decontamination and source-pool construction, using the decontaminated global cross statistics.
The source pool consists of high-quality medical samples, strong medical supplements from core biomedical labels, and high-quality general samples.
Although QTE remains the largest bucket under the natural source-pool distribution, the BVE-enhanced final mixture deliberately increases BVE to 45\% of tokens while retaining 30\% QTE, 20\% MSE, and 5\% AUX.

\begin{table}[t]
  \caption{Evidence-bucket distribution in the decontaminated source pool before final mixture sampling.}
  \label{tab:source-pool-buckets}
  \centering
  {\small
    \setlength{\tabcolsep}{1mm}
    \begin{tabular*}{\textwidth}{@{\extracolsep{\fill}} cccccc}
      \toprule
      \textbf{Evidence bucket} & \textbf{Samples} & \textbf{Text tokens} & \textbf{Image tokens} & \textbf{Total tokens} & \textbf{Share} \\
      \midrule
      Biomedical Visual Evidence (BVE) & 4,036,696 & 2.07B & 2.27B & 4.33B & 28.13\% \\
      Quantitative/Table Evidence (QTE) & 8,681,434 & 3.72B & 4.30B & 8.02B & 52.03\% \\
      Mechanism/Structure Evidence (MSE) & 3,117,352 & 1.18B & 1.37B & 2.55B & 16.54\% \\
      Auxiliary (AUX) & 621,590 & 0.21B & 0.30B & 0.51B & 3.31\% \\
      \midrule
      Total & 16,457,072 & 7.18B & 8.23B & 15.41B & 100.00\% \\
      \bottomrule
    \end{tabular*}
  }
\end{table}

Table~\ref{tab:final-mixture-buckets} reports the final PMC-InterCPT bucket composition after modality-aware resampling.
Compared with the natural source-pool distribution, the final mixture increases BVE from 28.13\% to 45.0\% while reducing QTE from 52.03\% to 30.0\%.
This produces a medical-visual-evidence-enhanced corpus while retaining substantial QTE and MSE coverage.

\begin{table}[h]
  \caption{Final evidence-bucket composition of PMC-InterCPT after modality-aware resampling.}
  \label{tab:final-mixture-buckets}
  \centering
  {\small
    \begin{tabularx}{\columnwidth}{
        Y
        Z{0.25\columnwidth}
        Z{0.25\columnwidth}
      }
      \toprule
      \textbf{Evidence bucket} & \textbf{Tokens} & \textbf{Final share} \\
      \midrule
      BVE                      & 4.33B           & 45.0\%               \\
      QTE                      & 2.89B           & 30.0\%               \\
      MSE                      & 1.93B           & 20.0\%               \\
      AUX                      & 0.48B           & 5.0\%                \\
      \midrule
      Total                    & 9.63B           & 100.0\%              \\
      \bottomrule
    \end{tabularx}
  }
\end{table}

Table~\ref{tab:pool-stats} gives the corresponding five-pool decomposition.
Pools 1--3 are included in the source pool used for all mixture experiments, while pools 4--5 are excluded from pretraining.

\begin{table}[h]
  \caption{Five quality/medical pools after decontamination. HQ denotes high-quality, and supp. denotes supplement. Pools 1--3 form the source pool, and pools 4--5 are excluded from pretraining.}
  \label{tab:pool-stats}
  \centering
  {\small
    \setlength{\tabcolsep}{3pt}
    \begin{tabularx}{\columnwidth}{
        Y
        Z{0.18\columnwidth}
        Z{0.12\columnwidth}
      }
      \toprule
      \textbf{Pool}               & \textbf{Samples} & \textbf{Tokens} \\
      \midrule
      Pool 1: HQ medical          & 12,389,828       & 11.67B          \\
      Pool 2: Med-strong supp.    & 782,139          & 1.02B           \\
      Pool 3: HQ general          & 3,285,105        & 2.73B           \\
      Pool 4: Med-weak supp.      & 2,278,176        & 2.61B           \\
      Pool 5: Low-quality general & 704,879          & 0.67B           \\
      \bottomrule
    \end{tabularx}
  }
\end{table}

\section{Sample Reconstruction and Filtering Details}
\label{app:sample-construction-details}

\subsection{Interleaved Sample Format}
Each reconstructed interleaved sample is serialized as a Parquet row with three top-level fields: \texttt{images}, \texttt{texts}, and \texttt{metadata}.
The \texttt{images} and \texttt{texts} fields are aligned arrays of equal length.
An image slot stores a base64-encoded image in \texttt{images} and \texttt{null} in the corresponding \texttt{texts} position. A text slot stores \texttt{null} in \texttt{images} and the text string in \texttt{texts}.
For a single-image sample, the typical sequence is:
\[
  \begin{aligned}
    [\mathrm{IMG}_1] & \rightarrow [\mathrm{Figure}\ n.\ \mathrm{caption}] \\
                     & \rightarrow [\mathrm{context\ paragraph}].
  \end{aligned}
\]
For multi-image samples, additional referenced figures are inserted with their captions before the shared context, so that the context can refer to all figures present in the sequence.

The \texttt{metadata} field is a JSON string that stores article-level provenance and bibliographic information together with aligned per-image attributes, including file names, identifiers, paths, captions, labels, panel annotations, cluster assignments, and image sizes.
Captions receive explicit figure-number prefixes, e.g., ``Figure~2.'', to disambiguate subsequent context references.

\subsection{Context Coherence Repair and Length Filtering}
Context coherence repair is applied after initial interleaved sample reconstruction.
For samples with multiple context segments, we locate each normalized context string in the full article text and check whether consecutive segments are adjacent in the original article.
Two segments are treated as adjacent when their spans are contiguous after whitespace normalization. The implementation also uses an edge-anchor fallback that searches for the concatenation of the previous segment's ending words and the next segment's beginning words in the article text.

When two consecutive contexts are not adjacent, we avoid placing unrelated statements next to one another in a synthetic context that may not faithfully represent the evidence associated with the figure.
The repair rule compares the two segments using raw figure-reference information from the source \texttt{image\_context}.
If only one segment refers exclusively to the primary image, that segment is kept.
If both segments are primary-image-only, or neither is primary-image-only, the earlier segment is kept.
After context pruning, images and captions that are no longer referenced by the kept contexts are removed, while the primary image is always retained.
The metadata arrays are subsetted consistently with the retained images.

Length filtering is applied after context cleanup.
For each row, the first caption and all non-caption context segments are counted separately.
The script removes tail figure/table-reference fragments from contexts, normalizes repeated whitespace, collapses duplicated periods, removes empty parentheses, and removes extra spaces before punctuation.
Language is determined by detecting CJK characters in the combined caption and context.
For Chinese samples, a row is dropped when both \texttt{caption\_chars < 40} and \texttt{context\_chars < 120}.
For non-Chinese samples, a row is dropped when both \texttt{caption\_tokens < 12} and \texttt{context\_tokens < 30}.

\section{Medical Relevance and Quality Classification}
\label{app:classification-details}

\subsection{Annotation Guidelines}
The annotator receives the captions and figure-linked context but not the images.
This separates medical relevance and text purity from figure-context association, which is supported by PMC figure-caption pairings and article-native references.
Appendix~\ref{app:teacher-quality-audit} compares this approach with image-conditioned scoring and illustrates what can go wrong when an alignment judgment also changes the quality score.

\paragraph{Medical relevance.}
\texttt{medical\_relevance} is binary.
A score of 1 indicates that the sample is medically, clinically, or biomedically relevant, including biomedical research, clinical imaging, pathology, laboratory assays, molecular biology, public health, pharmacology, and other life-science evidence.
A score of 0 indicates that the text is not medical or biomedical, or is only incidentally related to science without useful medical-domain content.

\paragraph{Text quality.}
\texttt{quality} is a three-level text-purity score.
The score does not measure whether the caption fully describes the image, because the annotator does not receive the image.
Instead, it measures whether the text is clean enough to serve as CPT data:

\begin{itemize}
  \item \textbf{0: severe quality problems.} The sample has heavy repetition, copy-paste artifacts, severe garbling, major incoherence, mixed unrelated passages, unreadable table dumps, malformed \LaTeX{} source fragments, or text dominated by extraction noise.
  \item \textbf{1: usable but imperfect.} The sample remains mostly understandable, but contains noticeable redundancy, corruption, malformed joins, table-formatting damage, citation clutter, or limited extraction artifacts.
  \item \textbf{2: clean and coherent.} The sample is low-noise, low-redundancy, readable medical or biomedical text with coherent captions and context.
\end{itemize}

The guidelines deliberately penalize flattened tables and malformed formula extraction.
Short, semantically meaningful equations or chemical notation are allowed, but raw source fragments such as \texttt{\textbackslash documentclass}, \texttt{\textbackslash usepackage}, long pasted \TeX{} blocks, duplicated rendered/source equations, and broken table rows lower the quality score.
If a table or formula artifact is present but limited, the maximum score is usually 1; if it dominates the sample or makes a large section difficult to read, the score is 0.

\subsection{LLM Annotation Prompt}
We use Gemini-3.1-pro-preview with temperature 0 to label sampled interleaved records.
The user message concatenates the scoring instruction with a text-only rendering of the sample.
The sample rendering separates captions from additional context so that the annotator can distinguish figure captions from body-text evidence.
The first block below shows the complete scoring instruction used in the annotation request.

\begin{promptbox}[System-style instruction used in the annotation request]
  \small
  You are grading one interleaved text-only sample for medical multimodal pretraining data filtering.

  \textbf{Important:}
  \begin{itemize}
    \item You will NOT receive images.
    \item Judge ONLY from the provided text.
    \item Do NOT assess completeness because images are unavailable.
    \item The final quality score is just a text purity score.
  \end{itemize}

  \textbf{You must score TWO dimensions:}
  \begin{enumerate}
    \item \texttt{medical\_relevance}: integer in \{0,1\}
    \item \texttt{quality}: integer in \{0,1,2\}
  \end{enumerate}

  \textbf{Definitions:}

  \textbf{1) Medical Relevance:}
  \begin{itemize}
    \item 0: not medical / not biomedical / not clinically relevant.
    \item 1: medically or biomedically relevant.
  \end{itemize}

  \textbf{2) Quality (text purity only):}
  \begin{itemize}
    \item 0: severe text quality problems. Examples: heavy repetition, obvious copy-paste artifacts, severe garbling, major incoherence, mixed unrelated content, unreadable table dumps, or text dominated by noise.
    \item 1: usable but imperfect. Some redundancy, corruption, noticeable noise, or table-related formatting damage remains.
    \item 2: clean, coherent, low-noise, low-redundancy medical text.
  \end{itemize}

  \textbf{Very strict penalty rules for tables and readability:}
  \begin{itemize}
    \item If the text contains a flattened table, embedded raw table dump, broken row/column listing, mashed cell values, or missing-space table extraction that noticeably hurts readability, quality must be lowered.
    \item If table artifacts are present but limited and the sample is still mostly readable prose, use quality=1, not 2.
    \item If table artifacts dominate a substantial portion of the sample, or make a large section hard to read, use quality=0.
    \item If the text switches from normal prose into malformed table content, this should usually be treated as at least quality=1 and often quality=0 if the malformed part is long or noisy.
    \item A clean sentence that merely references a table is not a problem. Only penalize when table content itself appears in the text and damages readability.
  \end{itemize}

  \textbf{Very strict penalty rules for \LaTeX{} / formula artifacts:}
  \begin{itemize}
    \item Clean, short, semantically normal mathematical expressions are acceptable and should NOT be penalized by themselves. Examples of acceptable content include inline or display math such as \texttt{\textbackslash(a\^{}2 - 4b = m\^{}2\textbackslash)}, chemical notation, or brief equations that naturally belong in scientific text.
    \item Penalize strongly when the text contains malformed, dumped, or extraction-corrupted \LaTeX{}/\TeX{} content that hurts readability. Examples include raw preamble or source fragments such as \texttt{\textbackslash documentclass}, \texttt{\textbackslash usepackage}, \texttt{\textbackslash begin\{document\}}, \texttt{\textbackslash end\{document\}}, duplicated rendered formula text, OCR-broken math markup, or long formula-source blocks pasted into prose.
    \item If a malformed \LaTeX{} block appears briefly but the rest of the sample is still understandable, use quality=1.
    \item If malformed \LaTeX{} or formula-source noise occupies a substantial part of the sample, interrupts sentence flow, or makes the sample feel corrupted or unnatural, use quality=0.
    \item When both a human-readable equation and a dumped \LaTeX{} source block appear together, treat this as a formatting artifact and lower quality.
  \end{itemize}

  \textbf{Other strict penalty rules for quality:}
  \begin{itemize}
    \item Repeated short sentences, repeated clauses, repeated local phrases, or repeated caption fragments should lower quality.
    \item Context that mostly repeats the caption rather than adding useful text should lower quality.
    \item Fragmented text, malformed joins, truncation artifacts, broken sentence boundaries, or mixed unrelated passages should lower quality.
    \item Citation-heavy but still readable text can be acceptable; only penalize if it materially hurts readability or purity.
    \item Be conservative: if unsure whether table artifacts materially reduce readability, choose the lower quality score.
  \end{itemize}

  \textbf{Output requirements:}
  \begin{itemize}
    \item Return ONLY one JSON object.
    \item No markdown, no extra keys.
  \end{itemize}

  \textbf{Return this exact schema:}

  {\ttfamily
    \{\\
    \quad "medical\_relevance": 0|1,\\
    \quad "quality": 0|1|2,\\
    \quad "reasons": \{\\
    \quad\quad "medical\_relevance": "one short sentence",\\
    \quad\quad "quality": "one short sentence"\\
    \quad \}\\
    \}
  }
\end{promptbox}

The second block shows the text-only sample template concatenated with the scoring instruction.

\begin{promptbox}[Text-only sample format]
  \texttt{Sample texts for evaluation:}

  \texttt{Captions:}

  \texttt{- <caption text for each image>}

  \texttt{Additional context texts:}

  \texttt{- <figure-linked body-text context>}
\end{promptbox}
\subsection{Classifier Training Details}
We train separate Qwen3-1.7B classifiers with LLaMA-Factory~\citep{zheng2024llamafactory}: a three-way text-quality classifier and a binary medical-relevance classifier.
Both use full-parameter fine-tuning with a 2,048-token cutoff, per-device batch size 4, no gradient accumulation, learning rate $1\times10^{-5}$, six epochs, cosine learning-rate decay, warmup ratio 0.1, and bf16 precision.
For each task, we retain the checkpoint with the strongest validation-set performance.

\begin{table}[h]
  \caption{Balanced split for the quality classifier. Labels 0, 1, and 2 correspond to severe-noise, usable-but-imperfect, and clean text quality.}
  \label{tab:quality-cls-split}
  \centering
  {\small
    \begin{tabularx}{\columnwidth}{
        Z{0.15\columnwidth}
        Y
        Y
        Y
        Y
      }
      \toprule
      \textbf{Split} & \textbf{Label 0} & \textbf{Label 1} & \textbf{Label 2} & \textbf{Total} \\
      \midrule
      Train          & 408              & 408              & 408              & 1,224          \\
      Validation     & 51               & 51               & 51               & 153            \\
      Test           & 52               & 52               & 52               & 156            \\
      \bottomrule
    \end{tabularx}
  }
\end{table}

\begin{table}[h]
  \caption{Balanced split for the medical-relevance classifier. Label 0 denotes non-medical and label 1 denotes medical.}
  \label{tab:medical-cls-split}
  \centering
  {\small
    \begin{tabularx}{\columnwidth}{
        Z{0.28\columnwidth}
        Y
        Y
        Y
      }
      \toprule
      \textbf{Split} & \textbf{Label 0} & \textbf{Label 1} & \textbf{Total} \\
      \midrule
      Train          & 1,134            & 1,134            & 2,268          \\
      Validation     & 141              & 141              & 282            \\
      Test           & 143              & 143              & 286            \\
      \bottomrule
    \end{tabularx}
  }
\end{table}

Large-scale classifier inference is run with SGLang~\citep{zheng2024sglang}.
Both classifiers use the same serving configuration: data parallelism 8, tensor parallel size 1, batch size 512, maximum input length 4,096 characters, maximum new tokens 4, and temperature 0.
The intended output is a single label per record. Invalid outputs count as errors in the medical-relevance evaluation.

\subsection{Classifier Test-Set Performance Analysis}
Table~\ref{tab:quality-cls-performance} and Table~\ref{tab:medical-cls-performance} report held-out test performance for the two classifiers.
The quality classifier reaches 92.3\% accuracy and 92.3\% macro F1.
Performance is strongest on label 0 and label 2, with F1 scores of 95.3\% and 93.2\%, respectively.
The lower F1 for label 1 (88.2\%) is expected because this middle class is a boundary category: limited table damage, mild repetition, or moderate formatting noise can be close to either severe-noise label 0 or clean label 2.
This error pattern is acceptable for filtering because the main data pool keeps quality$\geq$1 records. The most important operational distinction is separating severely corrupted samples from usable ones.

\begin{table}[h]
  \caption{Quality classifier performance on the held-out test set.}
  \label{tab:quality-cls-performance}
  \centering
  {\small
    \begin{tabularx}{\columnwidth}{
        Z{0.68\columnwidth}
        Y
      }
      \toprule
      \textbf{Metric} & \textbf{Value} \\
      \midrule
      Accuracy        & 92.3\%         \\
      Macro F1        & 92.3\%         \\
      Label 0 F1      & 95.3\%         \\
      Label 1 F1      & 88.2\%         \\
      Label 2 F1      & 93.2\%         \\
      \bottomrule
    \end{tabularx}
  }
\end{table}

Against the Gemini reference labels on the balanced held-out test set of 286 records, the medical-relevance classifier reaches 93.4\% accuracy (267/286) and 93.7\% macro F1.
Medical recall is 90.2\% (129/143), and medical precision is 96.3\% (129/134). Of the 143 medical reference records, 12 were predicted as non-medical and two produced invalid outputs. Among the 143 non-medical reference records, five were predicted as medical.

The 6.6\% error rate (19/286) counts false negatives, false positives, and invalid outputs against the Gemini reference labels on the balanced test set. Inputs classified as non-medical can still be used for training through the high-quality general pool if their quality score is $q=2$. Inputs classified as non-medical with $q=1$ do not enter the medical supplement.

\begin{table}[h]
  \caption{Medical-relevance classifier performance on the held-out test set.}
  \label{tab:medical-cls-performance}
  \centering
  {\small
    \begin{tabularx}{\columnwidth}{
        Z{0.68\columnwidth}
        Y
      }
      \toprule
      \textbf{Metric}      & \textbf{Value} \\
      \midrule
      Accuracy             & 93.4\%         \\
      Macro F1             & 93.7\%         \\
      Medical precision    & 96.3\%         \\
      Medical recall       & 90.2\%         \\
      Medical label F1     & 93.1\%         \\
      Non-medical label F1 & 94.2\%         \\
      \bottomrule
    \end{tabularx}
  }
\end{table}

\subsection{Medical-Relevance Audit by MeSH Topic Frequency}
\label{app:mesh-recall-audit}

\begin{figure}[htbp]
  \centering
  \includegraphics[width=\textwidth]{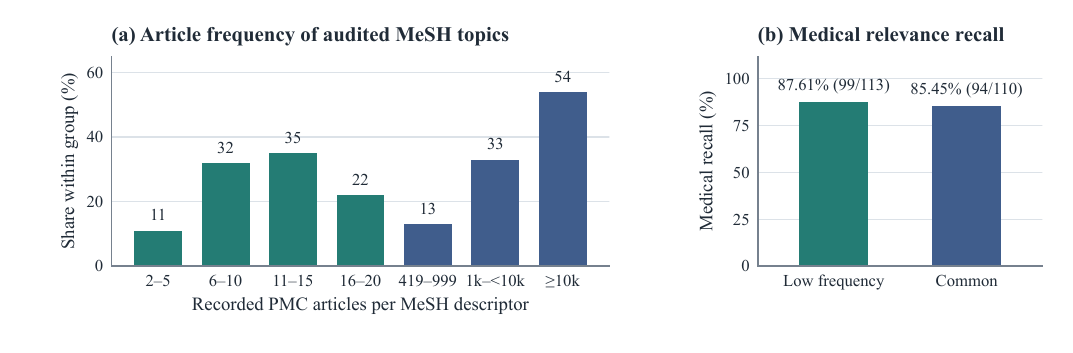}
  \caption{Medical-relevance audit by MeSH topic frequency. \textbf{(a)} Distribution of the recorded article frequencies of target MeSH topics across the 100 pairs in each group. \textbf{(b)} Medical-relevance recall in the paired sample. Counts above the bars show the number recalled out of all medically relevant inputs.}
  \label{fig:mesh-recall-audit}
\end{figure}

We compared the recall of our medical-relevance classifier on inputs sampled from low-frequency and common Medical Subject Headings (MeSH) topics. We measured topic frequency by the recorded number of pre-filter PMC articles associated with each MeSH descriptor. In the selected sample, the recorded counts ranged from 2 to 20 articles for low-frequency topics and from 419 to 104,919 for common topics (Figure~\ref{fig:mesh-recall-audit}a).

For each article, we paired a predicted medical input with evidence of the target MeSH topic and a predicted non-medical input. The low-frequency and common groups each contained 200 inputs, with 100 classified as medical and 100 as non-medical by our medical-relevance classifier. For the inputs we reviewed manually, we used human annotations as the medical-relevance reference labels.

In the paired sample, the classifier recalled 99 of 113 medically relevant inputs in the low-frequency group (\textbf{87.61\%}) and 94 of 110 in the common group (\textbf{85.45\%}), missing 14 and 16 inputs, respectively (Figure~\ref{fig:mesh-recall-audit}b). The classifier thus recovered most medically relevant inputs in both groups, including those drawn from low-frequency topics.

\subsection{Paired Audit of Teacher Annotations}
\label{app:teacher-quality-audit}

We sampled 200 interleaved records after reconstruction and before quality-based filtering. Both workflows used Gemini 3.1 Pro Preview to score figure captions and linked passages from the article body. The image-conditioned workflow also received the corresponding figures, and its rubric allowed clear image-text mismatches to lower the quality score. We manually annotated the quality of these samples and used the resulting labels as a reference. Against reference scores, exact agreement was 87.5\% (175/200) for the text-only workflow and 71.0\% (142/200) for the image-conditioned workflow.

\paragraph{Quality-score disagreements.}
In all 67 disagreements, the image-conditioned workflow assigned a score of 1. Among the 29 changes from 0 to 1, the text-only rationale cited raw \LaTeX{} in 24 cases, embedded SVG source in three, and malformed tables in two. The image-conditioned workflow often identified the same artifacts but deemed the sample usable because some prose remained readable. Among the 38 changes from 2 to 1, only three image-conditioned rationales explicitly cited a problem in the supplied figures. Some gave no specific reason for the downgrade; one stated that the content was clear and aligned with the figure but still assigned 1.

\paragraph{Why score calibration matters.}
A change from 0 to 1 can admit corrupted text past the initial quality filter. Severe extraction artifacts can replace coherent caption or context text with markup or flattened table entries, weakening the signal available for pretraining. A change from 2 to 1 can move a non-core medical record into the medical weak-related supplement, or a non-medical record into the low-quality general pool. Neither pool is used in evidence-aware allocation, so unnecessary downgrades can reduce the amount of data available to build the final corpus.

\begin{figure}[htbp]
  \centering
  \tcbset{audit case/.style={
    enhanced,
    frame hidden,
    colback=white,
    boxrule=0pt,
    arc=4pt,
    borderline={0.6pt}{0pt}{black!65,dashed},
    boxsep=0pt,
    left=6pt, right=6pt, top=6pt, bottom=6pt,
    before skip=0pt, after skip=0pt
  }}
  \begin{minipage}[t]{0.47\linewidth}
    \vspace{0pt}
    \begin{tcolorbox}[audit case]
      \centering
      \small\textbf{(a) Unsupported downgrade}\par\smallskip
      \includegraphics[width=\linewidth,height=0.28\textheight,keepaspectratio]{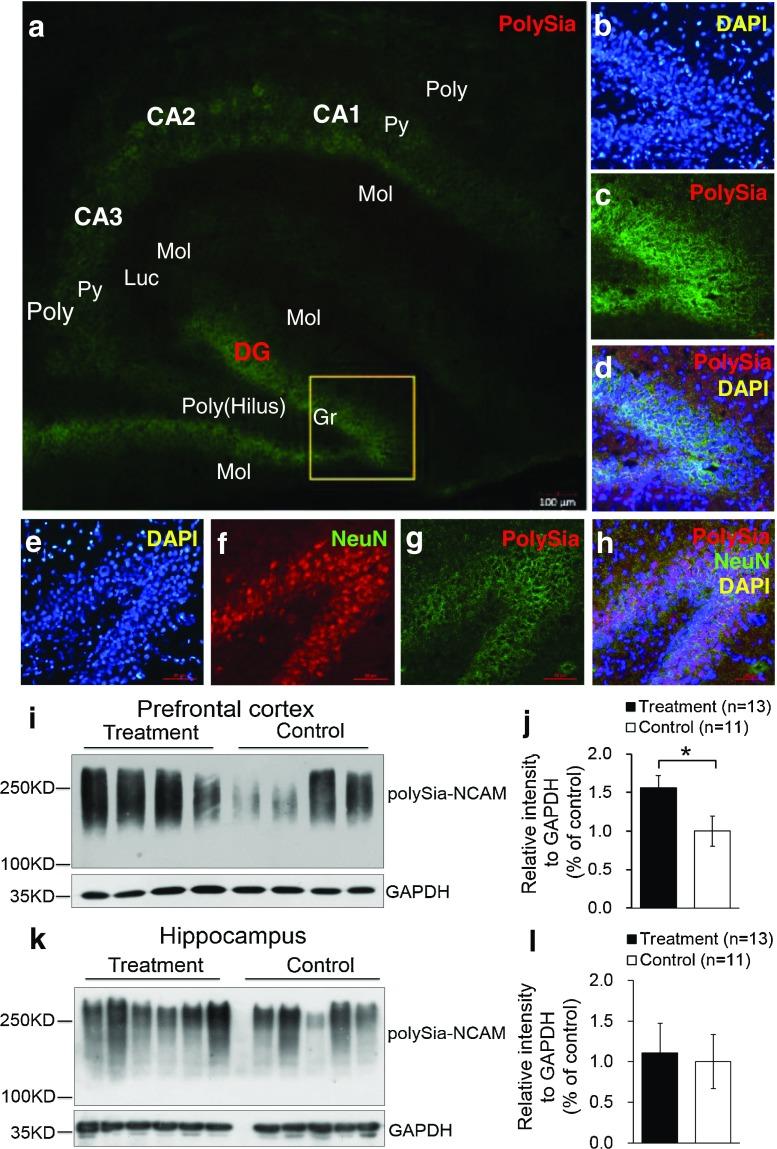}\par
      \smallskip
      \raggedright
      \textbf{End of the input context:}
      ``To our knowledge, these new findings have not been previously reported''\par
      \smallskip
      \textbf{Model rationale ($q=1$):}
      ``\ldots{} quality is reduced due to an appended image tag at the end.''
    \end{tcolorbox}
  \end{minipage}\hfill
  \begin{minipage}[t]{0.47\linewidth}
    \vspace{0pt}
    \begin{tcolorbox}[audit case]
      \centering
      \small\textbf{(b) Missed figure-text mismatch}\par\smallskip
      \includegraphics[width=\linewidth,height=0.28\textheight,keepaspectratio]{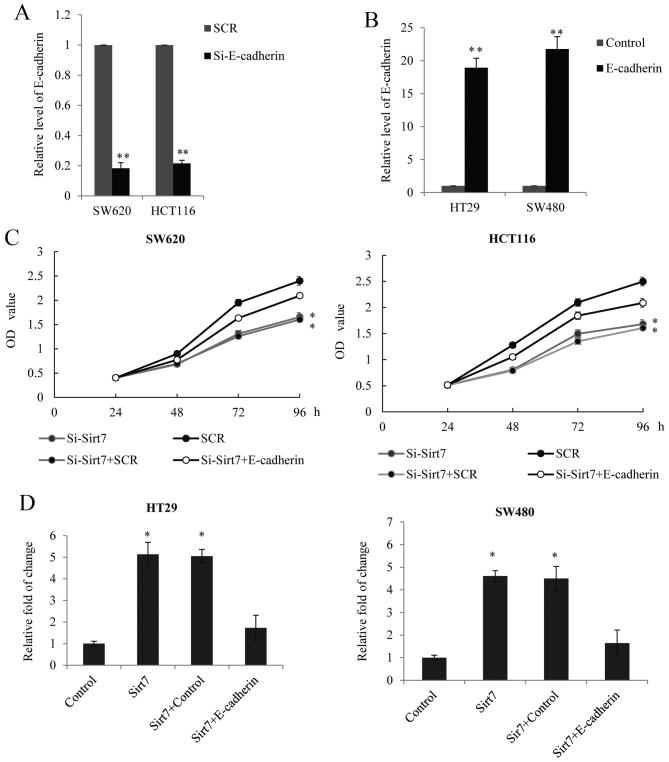}\par
      \smallskip
      \raggedright
      \textbf{Source caption excerpt:}
      ``Endogenous E-cadherin level was measured in \textbf{SW620} and \textbf{HCT116} cells''\par
      \smallskip
      \textbf{Panel B labels:} \textbf{HT29} and \textbf{SW480}.\par
      \smallskip
      \textbf{Model rationale ($q=2$):}
      ``\ldots{} matches the provided figure.''
    \end{tcolorbox}
  \end{minipage}
  \caption{Two failure cases in image-conditioned quality scoring. The supplied images are reproduced from Fig.~4 of \citet{chen2015lactoferrin} and Fig.~5 of \citet{deng2018sirtuin}, respectively.}
  \label{fig:r1w2-visual-audit}
\end{figure}

\paragraph{Illustrative errors.}
Figure~\ref{fig:r1w2-visual-audit} shows two failures of image-conditioned scoring. In (a), the model hallucinated an appended image tag as a reason to assign 1: the input passage ends with ordinary prose and contains no such tag. The caption and panel~j do disagree on the treatment group size ($n=15$ vs.\ $n=13$), but the model did not cite that discrepancy. In (b), the model assigned 2 and claimed the text matched the figure, overlooking a discrepancy already present in the source article: panel~B is labeled HT29 and SW480, whereas the caption and body name SW620 and HCT116.

\section{Qualitative Case Study}
\label{app:qualitative-case-study}
\begin{figure}[t]
  \centering
  \includegraphics[width=\textwidth]{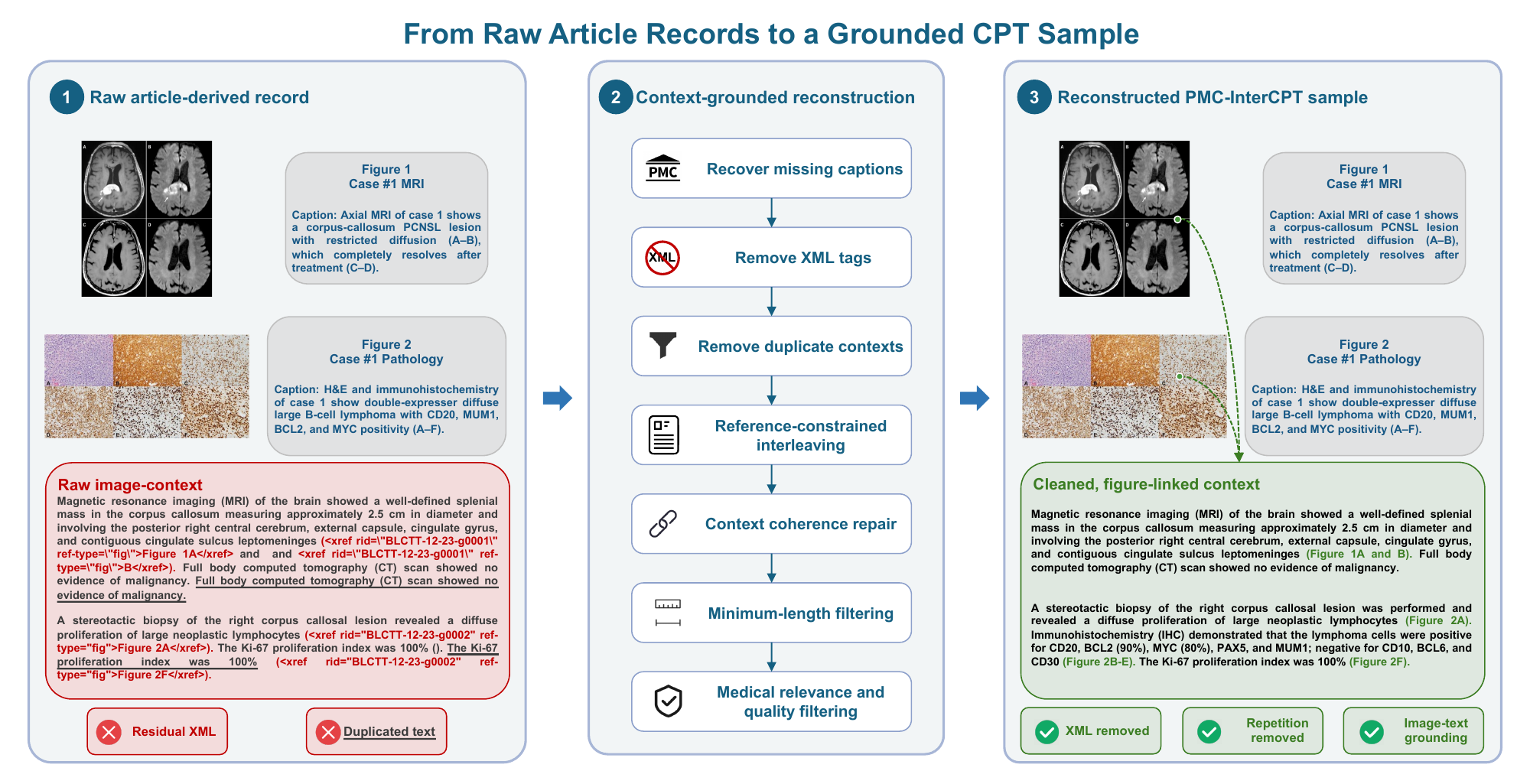}
  \caption{Illustrative reconstruction of a PMC-OA article record. The raw record contains residual \texttt{xref} markup and duplicated text in its image-context field. The reconstruction process removes these artifacts and preserves the jointly referenced MRI and pathology figures with normalized captions and a shared, figure-linked context.}
  \label{fig:reconstruction-case-study}
\end{figure}

Figure~\ref{fig:reconstruction-case-study} illustrates the transformation of one PMC-OA article-derived record. The raw image-context field contains residual \texttt{xref} markup and repeated text while referring jointly to an MRI figure and a pathology figure. Source-grounded recovery and reference-constrained interleaving retain the two figures with canonicalized captions and one cleaned shared context. Thus, the final sequence preserves the original article's cross-figure evidence while removing extraction artifacts, rather than generating new clinical content or attaching arbitrary nearby text.

\section{Evaluation Details}
\label{app:evaluation-details}

\subsection{Benchmark Descriptions}
MMMU~\citep{yue2024mmmu} is a multi-discipline multimodal reasoning benchmark covering college-level expert subjects.
We report MMMU-Med-Test using the Health and Medicine domain, and MMMU-All-Val using the full validation split across disciplines.
MMMU-Pro~\citep{yue2025mmmu} is a harder and more robust variant of MMMU with expanded option sets and reduced shortcut cues. We evaluate its medical subset with the 10-option configuration.

PMC-VQA~\citep{zhang2023pmcvqa} is a medical visual question answering benchmark built from PubMed Central figures.
We evaluate a cleaned test split, PMC-VQA-clean, and remove CPT samples whose source PMC articles overlap with this benchmark before training.
OmniMedVQA~\citep{hu2024omnimedvqa} is a broad medical VQA benchmark spanning multiple modalities and question categories, including clinical imaging, pathology, charts, and other biomedical visual evidence.
PretexEval~\citep{zhou2024pretexeval} evaluates medical knowledge mastery through predicate-equivalence transformations: it tests whether a model gives consistent answers to medically equivalent reformulations of a question.

ChartQA~\citep{masry2022chartqa} evaluates chart understanding through human-written and machine-augmented questions over chart images.
CharXiv~\citep{wang2024charxiv} evaluates reasoning over figures from scientific papers, with our experiments using the validation split's reasoning questions.
SciVQA~\citep{borisova2025scivqa} evaluates scientific figure question answering with a mixture of multiple-choice and open-ended questions.
Together, these general/scientific benchmarks test whether medical CPT improves biomedical reasoning without collapsing chart, scientific figure, and broad multimodal competence.

\subsection{Evaluation Protocol}
All benchmarks are evaluated with zero-shot prompting unless otherwise stated.
For multiple-choice questions, the prompt lists the question followed by options and instructs the model to answer directly with the option letter:

\begin{promptbox}[Multiple-choice evaluation prompt]
  \texttt{Question: <question>}\\
  \texttt{Options:}\\
  \texttt{A. <option A>}\\
  \texttt{B. <option B>}\\
  \texttt{...}\\
  \texttt{Answer with the option's letter from the given choices directly.}
\end{promptbox}

For open-ended questions, the prompt is:

\begin{promptbox}[Open-ended evaluation prompt]
  \texttt{<question>}\\
  \texttt{Answer the question using a single word or phrase.}
\end{promptbox}

Inference uses deterministic decoding with temperature 0.
The Medevalkit default configuration uses maximum new tokens 1024, top-$p$ 0.001, repetition penalty 1.0, presence penalty 2.0.

\paragraph{Rule-based scoring.}
MMMU and MMMU-Pro use exact option-letter accuracy.
PMC-VQA-clean and OmniMedVQA use rule-based multiple-choice matching against the gold option letter or answer text.
PretexEval uses rule-based answer matching.
No rule-based score is used for ChartQA, CharXiv-Val, or SciVQA in our reported results.

\begin{table}[t]
  \caption{Post-warmup maximum CPT learning-rate comparison for Qwen3.5-4B-Base on Natural-1B CPT with fixed SFT. Bold and underlining mark the best and second-best values.}
  \label{tab:lr-values}
  \centering
  {\scriptsize
    \setlength{\tabcolsep}{0.6mm}
    \begin{tabular*}{\textwidth}{@{\extracolsep{\fill}} Z{0.13\textwidth}*{11}{c}}
      \toprule
      \multirow{2}{*}{\textbf{\shortstack{Post-warmup\\max. CPT LR}}} & \multicolumn{6}{c}{\textbf{Medical benchmarks}} & \multicolumn{5}{c}{\textbf{General/scientific benchmarks}} \\
      \cmidrule(lr){2-7}\cmidrule(lr){8-12}
      & \textbf{\shortstack{MMMU\\Med}} & \textbf{\shortstack{MMMU-Pro\\Med}} & \textbf{\shortstack{PMC\\VQA}} & \textbf{\shortstack{Omni\\Med}} & \textbf{\shortstack{Pretex\\Eval}} & \textbf{\shortstack{Medical\\avg.}} & \textbf{\shortstack{MMMU\\All}} & \textbf{\shortstack{Chart\\QA}} & \textbf{\shortstack{Char\\Xiv}} & \textbf{\shortstack{Sci\\VQA}} & \textbf{\shortstack{General\\avg.}} \\
      \midrule
      $2.5\times10^{-6}$ & 59.30 & 39.86 & \textbf{61.85} & \underline{85.04} & 62.48 & 61.71 & \underline{54.44} & 76.44 & 50.40 & 66.74 & 62.01 \\
      $5\times10^{-6}$ & \textbf{60.10} & \underline{41.26} & \underline{61.10} & \textbf{86.38} & \underline{62.92} & \textbf{62.35} & \textbf{55.33} & \textbf{77.08} & \textbf{53.30} & \textbf{68.60} & \textbf{63.58} \\
      $1\times10^{-5}$ & \underline{59.42} & \textbf{41.96} & 60.80 & 84.45 & \textbf{63.84} & \underline{62.09} & \underline{54.44} & 76.20 & \underline{52.00} & \underline{67.29} & 62.48 \\
      $3\times10^{-5}$ & 57.71 & 37.76 & 60.70 & 84.67 & 61.10 & 60.39 & 54.33 & \textbf{77.08} & 51.40 & 67.26 & \underline{62.52} \\
      $5\times10^{-5}$ & 54.34 & 34.27 & 60.45 & 81.46 & 62.49 & 58.60 & 53.33 & \underline{76.48} & 47.20 & 65.26 & 60.57 \\
      \bottomrule
    \end{tabular*}
  }
\end{table}

\paragraph{LLM-judge scoring.}
For ChartQA, CharXiv-Val, and SciVQA, we use Qwen2.5-72B-Instruct as an LLM judge for all reported scores.
The judge receives the question, the standard answer, and the model response, and outputs a binary decision in the format \texttt{<judge>0</judge>} for correct or \texttt{<judge>1</judge>} for incorrect.
The judge prompt treats semantically equivalent answers as correct, which is important for open-ended scientific and chart questions where surface forms can differ from the reference.

\section{Additional Results}
\label{app:additional-results}

Unless otherwise noted, bold and underlining mark the best and second-best values.

\subsection{Full-Corpus Gains and Mixture Robustness}
\label{app:uncertainty}

\paragraph{Full-corpus gains at fixed checkpoints.}
We report 95\% confidence intervals for the difference between PMC-InterCPT and each baseline at the 4B and 2B scales. The medical average weights five benchmarks equally, and the general/scientific average weights four equally. The intervals quantify evaluation uncertainty at fixed checkpoints.

Table~\ref{tab:paired-difference-ci} shows that PMC-InterCPT improves both averages at both model scales relative to token-matched raw BIOMEDICA. The gains also hold against the larger, full raw corpus, with all eight intervals for the raw-data comparisons above zero. This supports the benefit of corpus construction beyond the amount of raw training data. Against SFT only, the 4B medical gain also excludes zero, while the other three intervals include zero.

\begin{table}[t]
  \caption{Score differences (percentage points) between PMC-InterCPT and each baseline at the 4B and 2B model scales. Brackets show 95\% confidence intervals.}
  \label{tab:paired-difference-ci}
  \vspace{3pt}
  \centering
  \small
  \begin{tabular*}{\linewidth}{@{\extracolsep{\fill}}llcc@{}}
    \toprule
    \textbf{Scale} & \textbf{Baseline} & \textbf{Medical avg.} & \textbf{General/scientific avg.} \\
    \midrule
    4B & SFT only & $+1.56\ [+0.54,\,+2.56]$ & $+0.07\ [-0.87,\,+1.02]$ \\
    4B & Full raw BIOMEDICA & $+1.58\ [+0.65,\,+2.52]$ & $+2.44\ [+1.54,\,+3.35]$ \\
    4B & Token-matched raw BIOMEDICA & $+1.46\ [+0.43,\,+2.32]$ & $+3.11\ [+2.11,\,+4.11]$ \\
    \midrule
    2B & SFT only & $+0.99\ [-0.16,\,+1.85]$ & $+0.45\ [-0.55,\,+1.43]$ \\
    2B & Full raw BIOMEDICA & $+1.53\ [+0.52,\,+2.51]$ & $+2.34\ [+1.34,\,+3.27]$ \\
    2B & Token-matched raw BIOMEDICA & $+2.36\ [+1.14,\,+3.17]$ & $+2.88\ [+1.57,\,+3.63]$ \\
    \bottomrule
  \end{tabular*}
\end{table}

\paragraph{Mixture choice across training seeds.}
We repeat the full CPT+SFT pipeline for Natural and BVE-enhanced under three matched seed groups, varying the seed in both training stages and keeping the CPT budget fixed at 1B tokens. Table~\ref{tab:sft-seed-variability} uses the same five medical benchmarks and four general/scientific benchmarks as the main results. BVE-enhanced improves the medical average in every seed group, with a mean paired gain of 1.01 points and a sample standard deviation of 0.27 points. The general/scientific difference changes sign and averages close to zero. These runs support selecting BVE-enhanced over Natural when prioritizing medical performance; they do not establish an optimal allocation.

\begin{table}[t]
  \caption{Natural and BVE-enhanced on Qwen3.5-4B-Base across three matched CPT+SFT seed groups. Scores are percentages. $\Delta$ is BVE-enhanced minus Natural in percentage points. The last row reports the mean and sample standard deviation across the three seed groups.}
  \label{tab:sft-seed-variability}
  \vspace{3pt}
  \centering
  \small
  \setlength{\tabcolsep}{2pt}
  \begin{tabular*}{\linewidth}{@{\extracolsep{\fill}}lcccccc@{}}
    \toprule
    & \multicolumn{3}{c}{\textbf{Medical average (5)}} & \multicolumn{3}{c}{\textbf{General/scientific average (4)}} \\
    \midrule
    \textbf{Seed group} & Natural & BVE-enhanced & $\Delta$ & Natural & BVE-enhanced & $\Delta$ \\
    \midrule
    G1 & 62.35 & 63.28 & $+0.92$ & 63.58 & 62.78 & $-0.80$ \\
    G2 & 61.24 & 62.55 & $+1.32$ & 62.69 & 62.86 & $+0.17$ \\
    G3 & 61.98 & 62.77 & $+0.79$ & 62.56 & 63.00 & $+0.44$ \\
    \midrule
    Mean $\pm$ SD & $61.85\pm0.57$ & $62.87\pm0.37$ & $+1.01\pm0.27$ & $62.94\pm0.56$ & $62.88\pm0.11$ & $-0.06\pm0.65$ \\
    \bottomrule
  \end{tabular*}
\end{table}

\subsection{Grounding Controls}

Table~\ref{tab:reference-breaking} reports results for the mixture-matched caption-only and reference-breaking controls. Breaking figure-context links lowers the medical average by 1.75 points and the general/scientific average by 3.40 points, indicating that the original figure-reference attachment contributes beyond in-domain article text alone.

\begin{table}[t]
  \caption{Figure-context controls on Qwen3.5-4B-Base. Superscripts give approximate CPT token budgets. The caption-only control retains images and captions. The reference-broken control replaces the original context with token-matched text from the same article. Bold marks the best score per column.}
  \label{tab:reference-breaking}
  \vspace{3pt}
  \centering
  {\fontsize{6}{7}\selectfont
    \renewcommand{\arraystretch}{1.08}
    \setlength{\tabcolsep}{0.75pt}
    \begin{tabular*}{\textwidth}{@{\extracolsep{\fill}}cc*{11}{c}}
      \toprule
      \textbf{Backbone} & \textbf{CPT condition} & \multicolumn{6}{c}{\textbf{Medical benchmarks}} & \multicolumn{5}{c}{\textbf{General/scientific benchmarks}} \\
      \cmidrule(lr){3-8}\cmidrule(lr){9-13}
      & & \textbf{\shortstack{MMMU\\Med}} & \textbf{\shortstack{MMMU-Pro\\Med}} & \textbf{\shortstack{PMC\\VQA}} & \textbf{\shortstack{Omni\\Med}} & \textbf{\shortstack{Pretex\\Eval}} & \textbf{\shortstack{Medical\\avg.}} & \textbf{\shortstack{MMMU\\All}} & \textbf{\shortstack{Chart\\QA}} & \textbf{\shortstack{Char\\Xiv}} & \textbf{\shortstack{Sci\\VQA}} & \textbf{\shortstack{General\\avg.}} \\
      \midrule
      \multirow{3}{*}{\shortstack{Qwen3.5-\\4B-Base}} & Reference-broken$^{1\mathrm{B}}$ & 56.39 & 39.86 & 61.40 & 85.17 & 64.80 & 61.52 & 53.33 & 77.44 & 46.60 & 60.14 & 59.38 \\
      & Caption-only$^{1\mathrm{B}}$ & 58.33 & 40.56 & 61.20 & 85.70 & 66.69 & 62.50 & 54.00 & \textbf{78.84} & 48.60 & 67.02 & 62.12 \\
      & PMC-InterCPT$^{1\mathrm{B}}$ & \textbf{59.36} & \textbf{41.96} & \textbf{61.50} & \textbf{86.19} & \textbf{67.37} & \textbf{63.28} & \textbf{55.44} & 76.64 & \textbf{51.60} & \textbf{67.45} & \textbf{62.78} \\
      \bottomrule
    \end{tabular*}
  }
\end{table}

\subsection{Learning-Rate Comparison Results}
\label{app:lr-values}

\begin{figure}[!htbp]
  \centering
  \includegraphics[width=0.55\textwidth]{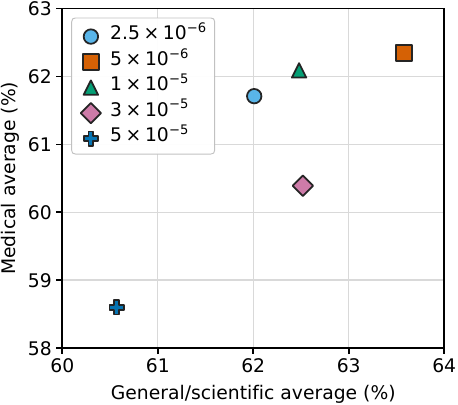}
  \caption{Aggregate medical and general results across post-warmup maximum CPT learning rates on Natural-1B with fixed SFT. Each color denotes one maximum learning rate.}
  \label{fig:cpt-lr-ablation}
\end{figure}

Figure~\ref{fig:cpt-lr-ablation} and Table~\ref{tab:lr-values} compare the maximum CPT learning rate reached after warmup on the Natural 1B mixture with fixed SFT. All entries use Qwen3.5-4B-Base and the fixed LLaVA-OneVision SFT stage. The $5\times10^{-6}$ configuration achieves the highest medical average (62.35) and general/scientific average (63.58), and is also best on MMMU-Med and OmniMedVQA. The neighboring $1\times10^{-5}$ configuration is strongest on MMMU-Pro Med and PretexEval, while $2.5\times10^{-6}$ is strongest on PMC-VQA-clean; neither improves both aggregate metrics. Raising the rate to $3\times10^{-5}$ and $5\times10^{-5}$ reduces both averages, especially at $5\times10^{-5}$.

\subsection{Per-Benchmark and Per-Subset Breakdown}
Unless otherwise stated, Tables~\ref{tab:additional-medical-breakdown}--\ref{tab:additional-scientific-breakdown} compare Qwen3.5-4B-Base after the same fixed SFT stage: raw BIOMEDICA uses 13.65B CPT tokens and PMC-InterCPT uses 9.63B CPT tokens.
Table~\ref{tab:additional-medical-breakdown} provides subject-level results for the two MMMU-style medical benchmarks.
PMC-InterCPT recovers much of the performance drop caused by raw BIOMEDICA CPT on the harder MMMU-Pro-Medical-10 setting, improving over BIOMEDICA by 2.80 points overall.
On MMMU-Pro-Med-10, the largest gains appear in Pharmacy (+8.77) and Public Health (+8.62), while Diagnostics and Laboratory Medicine improves over BIOMEDICA on both MMMU-Med-Test and MMMU-Pro-Med-10.
The gains are not uniform across all subjects, which is expected because the final mixture is optimized for broad medical multimodal CPT rather than subject-specific overfitting.

Table~\ref{tab:additional-omnimed-breakdown} reports OmniMedVQA results by question type and image modality.
PMC-InterCPT improves most clearly on Disease Diagnosis (+1.94 over BIOMEDICA) and Lesion Grading (+7.10), two categories that require clinically grounded visual interpretation rather than only recognizing the imaging modality.
The modality-level results show a similar pattern: PMC-InterCPT improves over BIOMEDICA on CT (+2.19), MR (+2.81), and Fundus Photography (+1.37), while gains are smaller or do not appear for several other modalities.
These category-level changes are consistent with the final curated, BVE-enhanced PMC-InterCPT improving clinically oriented visual questions, while not isolating the contribution of resampling from the other corpus-construction stages.

\begin{table}[t]
  \caption{Per-subset medical benchmark results for Qwen3.5-4B-Base. Raw BIOMEDICA uses 13.65B CPT tokens and PMC-InterCPT uses 9.63B; both use the same fixed SFT stage. Bold marks the best values.}
  \label{tab:additional-medical-breakdown}
  \centering
  {\scriptsize
    \setlength{\tabcolsep}{4pt}
    \begin{tabularx}{\textwidth}{
        Z{0.22\textwidth}
        Y
        Z{0.12\textwidth}
        Z{0.16\textwidth}
      }
      \toprule
      \textbf{Benchmark} & \textbf{Subset}                     & \textbf{BIOMEDICA} & \textbf{PMC-InterCPT} \\
      \midrule
      MMMU-Med           & Total                               & 58.62              & \textbf{58.73}        \\
      MMMU-Med           & Basic Medical Science               & 66.87              & \textbf{67.18}        \\
      MMMU-Med           & Clinical Medicine                   & \textbf{62.77}     & 61.23                 \\
      MMMU-Med           & Diagnostics and Laboratory Medicine & 46.91              & \textbf{48.77}        \\
      MMMU-Med           & Pharmacy                            & 60.00              & \textbf{63.49}        \\
      MMMU-Med           & Public Health                       & \textbf{53.24}     & 50.88                 \\
      \midrule
      MMMU-Pro-Med-10    & Total                               & 37.76              & \textbf{40.56}        \\
      MMMU-Pro-Med-10    & Basic Medical Science               & \textbf{50.00}     & 46.15                 \\
      MMMU-Pro-Med-10    & Clinical Medicine                   & \textbf{35.59}     & 32.20                 \\
      MMMU-Pro-Med-10    & Diagnostics and Laboratory Medicine & 23.33              & \textbf{26.67}        \\
      MMMU-Pro-Med-10    & Pharmacy                            & 49.12              & \textbf{57.89}        \\
      MMMU-Pro-Med-10    & Public Health                       & 32.76              & \textbf{41.38}        \\
      \bottomrule
    \end{tabularx}
  }
\end{table}

\begin{table}[t]
  \caption{OmniMedVQA breakdown by question type and image modality for Qwen3.5-4B-Base. Bold marks the best values.}
  \label{tab:additional-omnimed-breakdown}
  \centering
  {\scriptsize
    \setlength{\tabcolsep}{4pt}
    \begin{tabularx}{\textwidth}{
        Z{0.13\textwidth}
        Y
        Z{0.10\textwidth}
        Z{0.12\textwidth}
        Z{0.13\textwidth}
      }
      \toprule
      \textbf{Group} & \textbf{Subset}             & \textbf{N} & \textbf{BIOMEDICA} & \textbf{PMC-InterCPT} \\
      \midrule
      Question       & Modality Recognition        & 11,565     & 98.66              & \textbf{98.69}        \\
      Question       & Disease Diagnosis           & 55,387     & 84.27              & \textbf{86.21}        \\
      Question       & Anatomy Identification      & 16,448     & 80.74              & \textbf{81.36}        \\
      Question       & Lesion Grading              & 2,098      & 53.00              & \textbf{60.10}        \\
      Question       & Other Biological Attributes & 3,498      & 89.88              & \textbf{91.08}        \\
      \midrule
      Modality       & Fundus Photography          & 5,398      & 79.49              & \textbf{80.86}        \\
      Modality       & Microscopy Images           & 5,680      & \textbf{84.91}     & 84.37                 \\
      Modality       & X-Ray                       & 7,916      & \textbf{87.44}     & 87.41                 \\
      Modality       & Dermoscopy                  & 6,679      & \textbf{83.74}     & 82.45                 \\
      Modality       & CT                          & 15,809     & 83.90              & \textbf{86.09}        \\
      Modality       & Ultrasound                  & 10,991     & 80.47              & \textbf{82.08}        \\
      Modality       & MR                          & 31,877     & 86.07              & \textbf{88.88}        \\
      Modality       & OCT                         & 4,646      & \textbf{95.72}     & 95.61                 \\
      \bottomrule
    \end{tabularx}
  }
\end{table}

\begin{table}[!b]
  \caption{Scientific and chart-oriented benchmark breakdowns for Qwen3.5-4B-Base. Bold marks the best values.}
  \label{tab:additional-scientific-breakdown}
  \centering
  {\scriptsize
    \setlength{\tabcolsep}{4pt}
    \begin{tabularx}{\textwidth}{
        Z{0.18\textwidth}
        Y
        Z{0.12\textwidth}
        Z{0.14\textwidth}
        Z{0.16\textwidth}
      }
      \toprule
      \textbf{Benchmark} & \textbf{Subset} & \textbf{N} & \textbf{BIOMEDICA} & \textbf{PMC-InterCPT} \\
      \midrule
      SciVQA             & Total           & 4,200      & 66.74              & \textbf{67.29}        \\
      SciVQA             & Multiple choice & 1,200      & \textbf{70.25}     & 69.17                 \\
      SciVQA             & Open-ended      & 3,000      & 65.33              & \textbf{66.53}        \\
      \midrule
      ChartQA            & Total           & 2,500      & 75.00              & \textbf{77.00}        \\
      ChartQA            & Human           & 1,250      & 63.76              & \textbf{67.28}        \\
      ChartQA            & Augmented       & 1,250      & 86.24              & \textbf{86.72}        \\
      \midrule
      CharXiv-Val        & Total           & 1,000      & 47.90              & \textbf{54.00}        \\
      CharXiv-Val        & Answer type 1   & 440        & 49.55              & \textbf{55.68}        \\
      CharXiv-Val        & Answer type 2   & 99         & 63.64              & \textbf{73.74}        \\
      CharXiv-Val        & Answer type 3   & 232        & 51.29              & \textbf{57.33}        \\
      CharXiv-Val        & Answer type 4   & 229        & 34.50              & \textbf{38.86}        \\
      \bottomrule
    \end{tabularx}
  }
\end{table}

Table~\ref{tab:additional-scientific-breakdown} gives additional breakdowns for the scientific and chart-oriented benchmarks.
PMC-InterCPT improves ChartQA mainly on the human-written split (+3.52), while the augmented split changes only slightly.
On SciVQA, PMC-InterCPT improves open-ended questions (+1.20) but slightly decreases multiple-choice accuracy (-1.08), producing a modest overall gain.
The largest scientific-figure improvement appears on CharXiv-Val, where PMC-InterCPT improves every answer-type group and raises the total score by 6.10 points.

\end{document}